\documentclass{article}

\usepackage[preprint]{neurips_2025}

\usepackage{makecell}
\usepackage{multirow}
\usepackage{algorithm}
\usepackage{float}
\usepackage[utf8]{inputenc} 
\usepackage[T1]{fontenc}    
\usepackage{hyperref}       
\usepackage{url}            
\usepackage{booktabs}       
\usepackage{amsfonts}       
\usepackage{nicefrac}       
\usepackage{microtype}      
\usepackage{xcolor}         
\usepackage{amsmath} 
\usepackage{array}
\usepackage{graphicx}
\usepackage{subcaption}
\usepackage{caption}
\usepackage{natbib}           
\title{M-learner:A Flexible And Powerful Framework To Study Heterogeneous Treatment Effect In Mediation Model}

\author{%
Xingyu Li\\
Department of Biostatistics, The University of Texas MD Anderson Cancer Center\\
\And
Qing Liu\\
Center for Design and Analysis, Amgen\\
\And
Tony Jiang\\
Center for Design and Analysis, Amgen\\
\And
Hong Amy Xia\\
Center for Design and Analysis, Amgen\\
\And
Brian P. Hobbs\\
Telperian\\
\And
Peng Wei\\
Department of Biostatistics, The University of Texas MD Anderson Cancer Center\\
}

\begin{document}

\maketitle

\begin{abstract}
We propose a novel method, termed the M-learner, for estimating heterogeneous indirect and total treatment effects and identifying relevant subgroups within a mediation framework. The procedure comprises four key steps. First, we compute individual-level conditional average indirect/total treatment effect Second, we construct a distance matrix based on pairwise differences. Third, we apply t-SNE to project this matrix into a low-dimensional Euclidean space, followed by K-means clustering to identify subgroup structures. Finally, we calibrate and refine the clusters using a threshold-based procedure to determine the optimal configuration. To the best of our knowledge, This is the first method capable of revealing the complex relationships among X, M, and Y within mediation analysis while effectively controlling the Type I error rate. 
Experimental results validate the robustness and effectiveness of the proposed framework. Application to the real-world Jobs II dataset highlights the broad adaptability and potential applicability of our method. Code is available at \url{https:
	//anonymous.4open.science/r/M-learner-C4BB}.
\end{abstract}

\section{Introduction}
 Randomized clinical trials are often costly and time-consuming, with significant delays between treatment administration and the observation of final outcomes such as the response variable \(Y\). However, mediators can serve as early indicators of treatment efficacy. For example, in colorectal cancer trials, cessation of tumor growth may act as a proxy for positive treatment response. This enables interim evaluation of treatment efficacy and the potential to adapt treatment strategies mid-trial based on changes in the mediator. In practice, treatment effects may be mediated only in a subset of patients. Furthermore, the chosen mediator may not be valid, or treatment effects may be homogeneous (i.e., uniformly effective or ineffective across individuals). Existing approaches—such as the T-learner \cite{kunzel2019metalearners}and Causal Random Forest\cite{athey2016recursive}have focused on heterogeneous treatment effect estimation but do not account for mediation mechanisms.

For heterogeneous total treatment effects,
\citet{foster2011subgroup} introduced the Virtual Twins to estimate heterogeneous total treatment effects with RCT data. Several methodological approaches within observational studies enable the estimation of flexible and accurate models of heterogeneous total treatment effects. \citet{shalit2017estimating,shi2019adapting,johansson2016learning,hassanpour2019learning} use neural networks to estimate heterogeneous total treatment effects.
\citet{athey2016recursive,wager2018estimation,athey2019estimating} use random forests to estimate heterogeneous total treatment effects.
\citet{kunzel2019metalearners} proposed the Meta-learners which consist of S-learner, T-learner, and X-learner. \citet{nie2021quasi} proposed R-learner, \citet{zhang2022towards} extented the R-learner to continuous treatment scenario. And other methods have been proposed, such as DR-learner and Lp-R-learner \citep{kennedy2023towards}. \citet{dwivedi2020stable} proposed model selection solution in causal inference. \citet{kim2024hierarchical} proposed the causal clustering method. 

When a mediator is present, the primary analytical focus is on evaluating the indirect treatment effect on the outcome via the mediator. These method decompose the average treatment effect into direct treatment effect and indirect effect\citep{lin1997estimating,preacher2015advances,robins1992identifiability,petersen2006estimation,van2008direct,imai2010general,tchetgen2012semiparametric,vanderweele2015explanation,vandenberghe2017boosting, dorresteijn2011estimating,pearl2022direct,vanderweele2009conceptual,vansteelandt2012natural,angrist2004treatment,imbens2004nonparametric}.
Recently, to address commonly observed intermediated confounders that would be affected by the covariates and then affect both mediators and outcome, multiple methods have been developed to extend the classical metiation analysis\citep{tchetgen2014identification,diaz2021nonparametric,diaz2023efficient,gilbert2024identification}. \citet{ge2023reinforcement,ge2025review}, \citet{luo2025multivariate} and \citet{wang2025dynamic}proposed the method to use reinforcement learning to deal with the dynamic mediation analysis.
\citet{cheng2022causal} use deep learning to estimate causal effects in mediation model.
However, these methods do not account for treatment effect heterogeneity in mediation model. Recently, \citet{ting2025estimating} proposed the method to use BART to estimate the heterogeneous mediation effect. This method is primarily designed for estimating heterogeneous treatment effects and does not provide an effective solution for determining the existence of heterogeneity or for identifying heterogeneous subgroups.
In contrast, our method is the first to incorporate treatment effect heterogeneity within a mediation framework and to provide an approach for detecting the existence of heterogeneous subgroups as well as heterogeneous regions. It can effectively identify heterogeneity and the corresponding region. This enables the identification of subgroups characterized by distinct indirect treatment effects and provides a principled way to evaluate the effectiveness of the mediator and causal decision making.

To bridge these gaps, we propose a flexible and powerful method, termed M-learner, designed to capture heterogeneity in  treatment effects mediated by a mediator, we aim to estimate the Conditional Average Total Treatment Effect (CATTE) and the Conditional Average Indirect Treatment Effect (CAITE) from randomized clinical trial (RCT) data, and to identify subpopulations that benefit from treatment by examining heterogeneity in these effects.

Our approach facilitates the identification of subgroups exhibiting differential response patterns through mediated pathways, thereby improving the interpretability and clinical relevance of treatment effect estimation.
The key contributions of this paper are:
\begin{itemize}
\item[1.] We propose M-learner, a flexible and powerful method for estimating the conditional indirect treatment effect, which operates independently of any specific model architecture.
\item[2.]We introduce a novel clustering approach for subgroup identification based on treatment effect heterogeneity. By leveraging differences in treatment effects, our method reformulates the unsupervised task of discovering benefiting subgroups into a supervised learning problem.
\item[3.]We propose a novel calibration framework to assess the effectiveness of mediators and evaluate the reliability of identified subgroups.
\end{itemize}

\section{Models}
\subsection{Model Formulation}
\begin{figure}
\centerline{\includegraphics[width=1\textwidth]{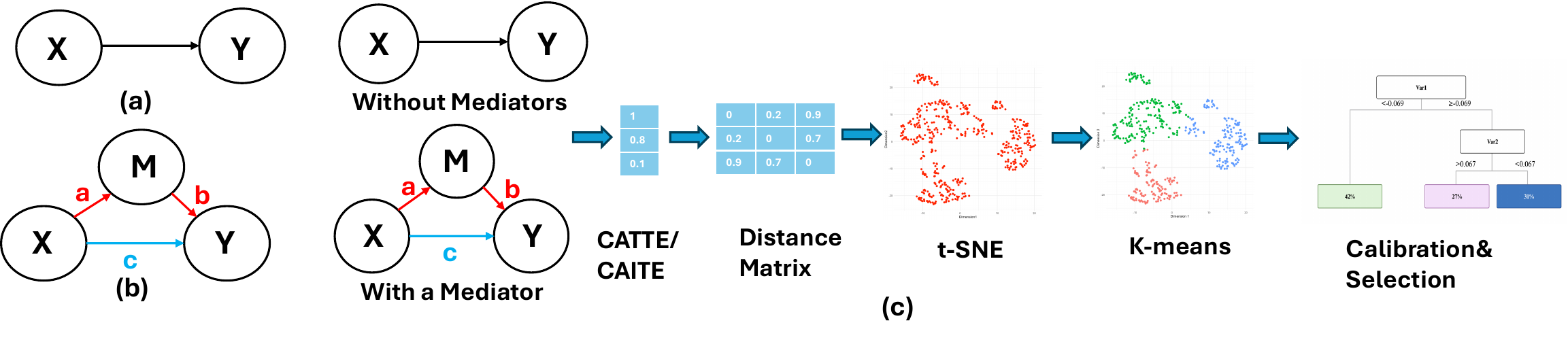}}
		\caption{(a)Directed Acyclic Graph between the covariates \(X\) and response \(Y\), (b)Directed Acyclic Graph between the mediator \(M\), covariates \(X\) and response \(Y\),(c)Pipeline of the M-learner.}
		\label{DAG}
\end{figure}

We assume the superpopulation or distribution \(\mathcal{P}\) from which  a realization of $N$ independent random variables is given as the training data. That is, \[(Y_i(0,M_i(0)),Y_i(1,M_i(0)),Y_i(0,M_i(1)),Y_i(1,M_i(1)),X_i,W_i,M_i(1),M_i(0))\sim \mathcal{P},\]where \(X_i\in\mathbb{R}^d\) is a d-dimensional covariate or feature vector, \(W_i\in\{0,1\}\) is the treatment-assignment indicator,\(Y_i(0, M_i(0))\in\mathbb{R}\) is the potential outcome of unit \(i\) when \(i\) is assigned to control group,  and the mediator \(M\) is assigned to control group, \(Y_i(1, M_i(0))\in\mathbb{R}\) is the potential outcome of unit \(i\) when \(i\) is assigned to control group,  and the mediator \(M\) is assigned to treatment group, \(Y_i(1, M_i(0))\in\mathbb{R}\) is the potential outcome of unit \(i\) when \(i\) is assigned to treatment group,  while the mediator \(M\) is assigned to control group,\(Y_i(1, M_i(1))\in\mathbb{R}\) is the potential outcome of unit \(i\) when \(i\) is assigned to treatment group,  and the mediator \(M\) is assigned to treatment group. What's more, this paper follows the four assumptions specified in the \citet{vanderweele2009conceptual} regarding the interrelations among covariate \(X\), outcome \(Y\), and the mediator \(M\).
When a mediator 
\(M\) is present, the relationships among 
\(X\), 
\(Y\), and 
\(M\) follow the structure shown in the  Directed Acyclic Graph (DAG) in Figure \ref{DAG} (b). The treatment effect can be transmitted through two distinct pathways. If the treatment effect is transmitted along path \(a\rightarrow b\)
via the mediator 
\(M\), this is referred to as the indirect treatment effect (ITE). If the treatment effect is transmitted directly to 
\(Y\)
along path 
\(c\), we refer to it as the direct treatment effect (DTE). The total treatment effect (TTE) is the combined influence of both the direct and indirect pathways. These effects are formally defined as follows,
\begin{eqnarray*}
\label{conception}
ITE&=&Y(1,M(1)) - Y(1,M(0)),DTE=Y(1,M(0)) - Y(0,M(0)),\nonumber\\
TTE&=&Y(1,M(1)) - Y(0,M(0))=ITE+DTE.
\end{eqnarray*}
When the mediator 
\(M\) is absent, the relationship between 
\(X\) and 
\(Y\) corresponds to the DAG in Figure \ref{DAG} (a). In this case, the treatment effect is transmitted to 
\(Y\) through a single pathway, representing the total treatment effect only.


For a new unit \(i\) with covariate vector \(x_i\), to decide whether to give the treatment via total treatment effect, we wish to estimate the individual 
total treatment effect (ITTE) of each \(i,D_i\), which is defined as 
    \(D_i := Y_i(1,M(1)) - Y_i(0,M(0)).\)
However, we can not observe \(D_i\) for any unit, and \(D_i\) is not identifiable without strong assumptions. Instead, \cite{kunzel2019metalearners} proposed that  we can estimate the CATTE function, which is defined as 
\(    \tau^{TOT}(x) = \mathbb{E}[D|X=x]=\mathbb{E}[Y(1,M(1) - Y(0,M(0)))|X=x],\)
the best estimator for the CATTE is also the best estimator for the ITTE in terms of the mean squared error (MSE). Let \(\hat{\tau}_i^{TOT}\) be an estimator of \(D_i\) and decompose the MSE at \(x_i\),
\begin{eqnarray}
\label{best_esti}
    \mathbb{E}[(D_i - \hat{\tau}_i^{TOT})^2|X=x_i] &=& \mathbb{E}[(D_i - \tau^{TOT}(x_i))^2|X=x_i]+\mathbb{E}[(\tau^{TOT}(x_i) - \hat{\tau}_i^{TOT})^2]. 
\end{eqnarray}
Because we can not affect the first term in \ref{best_esti}, the estimator that minimizes the MSE for the ITTE of \(i\) also minimizes the MSE for the CATTE at \(x_i\). To decide whether to give the treatment via indirect treatment
effect, we wish to estimate the individual indirect treatment effect (IITE) of each \(i\), where the definition of IITE is \(Y_i(1,M(1)) - Y_i(1,M(0))\). Analogous to Expression \ref{best_esti}, the CAITE can also be regarded as the optimal estimator of the IITE under the MSE criterion, where the definition of CAITE is \(\tau^{ITE}:=\mathbb{E}[Y(1,M(1)) - Y(1,M(0))|X].\) 
With the estimation of CATTE and CAITE, we can use the estimation to study the subgroups of TTE and ITE. For subgroups via TTE, it can be defined as 
\(
    \mathbb{E}[Y(1,M(1)) - Y(0,M(0))\in \mathcal{U}_i^{TTE}|X],i=1,2,3,\cdots,
\)
where \(\mathcal{U}_i^{TTE}\) is the set, the number of the sets are unknown, and when there is no heterogeneity, the number of the set is 1. For subgroups via ITE, it can be defined as 
\(
    \mathbb{E}[Y(1,M(1)) - Y(1,M(0))\in \mathcal{U}_i^{ITE}|X],i=1,2,3,\cdots,
\)
where \(\mathcal{U}_i^{ITE}\) is the set, the number of the sets are unknown, and when there is no heterogeneity, the number of the set is 1.

We propose a machine learning–based approach to separately estimate the CATTE and the CAITE, with the goal of identifying subgroups characterized by differences in both total and indirect treatment effects. This study addresses several key research questions: How can treatment effects be accurately estimated? How can meaningful and interpretable subgroups be identified based on treatment effect heterogeneity? Are the resulting subgroups statistically valid and practically useful? Finally, to what extent is the mediator effective—that is, does a significant portion of the treatment effect operate through the mediator?




\subsection{M-learner}

We propose an algorithm called M-learner\footnotemark shown in Figure \ref{DAG} (c) to address the aforementioned problem. This algorithm consists of four main steps:
\footnotetext{We refer to the method as the M-learner, as it is specifically designed to operate within the framework of mediation models.}
\begin{itemize}
    \item [1.]Estimating individual treatment effect (either the total treatment effect or the indirect treatment effect),
    \item [2.]Measuring the difference in treatment effects between pairs of samples (referred to as the “distance”), 
    \item [3.]Projecting the distance matrix into a two-dimensional Euclidean space using t-SNE, and clustering the projected data use K-means,
    \item [4.]Selecting the optimal clustering result and calibration.
\end{itemize}

When studying treatment effects within models involving mediators, we typically distinguish between the total treatment effect, which does not account for the mediator, and the indirect treatment effect transmitted through the mediator. In the following sections, we systematically discuss methods applicable to different scenarios and propose corresponding solutions.

%
In the absence of the mediator, the first step of the algorithm estimates the ITTE. At this stage, the T-learner is employed to estimate the CATTE. First, the control reponse function,
\({g}_0(x) = \mathbb{E}[Y(0)|X=x],\)
is estimated by a base learner, which could be any supervised learning estimator using the observations in the control group, \(\{(X_i,Y_i)\}_{W_i=0}\)and denoting the estimator by \(\hat{g}_0\). Second,  the treatment response function,
    \({g}_1(x) = \mathbb{E}[Y(1)|X=x],\) is estimated
with a base learner, using the treated observations \(\{(X_i,Y_i)\}_{W_i=1}\) and denoting the estimator by \(\hat{g}_1\). The T-learner is then obtained as 
    \(\hat{\tau}^{tot}(x) = \hat{g}_1(x) - \hat{g}_0(x).\)

In the presence of a mediator, the first step of the algorithm estimates the IITE. we use M-learner proposed in our paper to estimate CAITE, which takes three steps.
First, estimate the treatment mediator function,
    \(g^{M}_1(x) = \mathbb{E}[M(1)|X=x],\)
with a base learner, using the treated observations \(\{(M_i,Y_i)\}_{W_i=1}\)and denoting the estimator by \(\hat{g}^M_1\). Second, estimate the control mediator function,
   \( g^{M}_0(x) = \mathbb{E}[M(0)|X=x],\)
with a base learner, using the treated observations \(\{(X_i,Y_i)\}_{W_i=0}\)and denoting the estimator by \(\hat{g}^M_0\). 
Third, fit the treatment response function,
    \(g^{Y}_1(x,m) = \mathbb{E}[Y(1,M(1))|X=x,M=m],\)
with a base learner, using the treated observations \(\{(X_i,M_i,Y_i)\}_{W_i=1}\)and denoting the estimator by \(\hat{g}^{Y}_1\).
So, the M-learner is then obtained as 
\(
    \hat{\tau}^{ITE}(x) = \hat{g}^{Y}_1(x, \hat{g}^M_1(x)) - \hat{g}^{Y}_1(x, \hat{g}^M_0(x)).\)
Now, We have estimated the CATTE and CAITE, how to get the subgroups of different units? We propose a new method to transform the estimation of treatment effects to the clustering.

As the algorithms used to identify subgroups are identical for both CATTE and CAITE scenarios, we henceforth refer to them collectively as treatment effects (TE). The estimated TE for each unit \(i\) is denoted by \(\hat{\tau}_i\).

First, we evaluate the treatment effect difference between each pair of units \(i\) and \(j\), which we refer to as the distance. It is defined by \(dis(i,j)\), The distance metric can be defined as Euclidean distance, Manhattan distance, or other formulations. In the following analysis, we consistently adopt Euclidean distance to compute the distance between units \(i\) and \(j\) which means \(dis(i,j)= (\hat{\tau}_i - \hat{\tau}_j)^2\).
Consequently, an \(n\times n\) distance matrix is obtained. This matrix is then projected into a two-dimensional Euclidean space using t-SNE\citep{van2008visualizing}, where each unit \(i\) is assigned a coordinate. Subsequently, K-means clustering is performed on the projected points in the Euclidean space. The range of cluster numbers (from 2 to \(k\)) for K-means clustering must be specified in advance, based on the sample size and the number of covariates; an excessively large \(k\) can result in substantial overfitting, based on our experience, we recommend \(k = \left\lfloor\sqrt{d}\right\rfloor+2\), where \(\lfloor\rfloor\) represents rounding down to the nearest integer.
Then, a decision tree is employed to model the clustering results obtained for each predefined number of clusters, with the objective of mapping the unknown categories to interpretable categorical information.
For each leaf of the decision tree, we refer to it as a subtype. We use \(p_{leaf}\)to select the final, unique subtype grouping from different decision tree results. The definition of 
\(p_{leaf}\)is as follows,
            \begin{eqnarray}
            M &=& \beta_1 \text{leaf} + \beta_2 W,\label{lm1}\\
			M &=& \beta_3 \text{leaf}+\beta_4 W+\beta_5 			\text{leaf}*W, \label{lm2}
		\end{eqnarray}
        the likelihood functions of \ref{lm1} and \ref{lm2} are \(L_0\) and \(L_1\), respectively. Then \(2(\log L_1 - \log L_0)\) follows a chi-squared distribution,where the degrees of freedom correspond to the number of decision tree leaves minus \(1\). Based on this, we calculate \(p_{leaf}\). The decision tree result with the minimal \(p_{leaf}\) is chosen as the final subtype classification. When no mediating variable \(M\) is involved, the corresponding expression can be reformulated by substituting 
\(M\) with the outcome variable 
\(Y\).

\section{Simulation}

In real-world datasets, ground truth causal treatment effects are rarely directly observable. Consequently, empirical evaluation of causal inference methods often relies on synthetic data. For such evaluations to yield meaningful conclusions, the synthetic data must closely reflect real-world characteristics. In all experiments, we employed Random Forests(RF) and XGBoost(XGB) as base learners\citep{breiman2001random,chen2016xgboost}. Unless otherwise stated, all experiments in this section are conducted with a sample size of 1000 and 10 covariates.
%
\subsection{Subgroup analysis without mediators}
To identify subgroups that benefit from treatment through total treatment effects, we design four scenarios that reflect varying real-world heterogeneity structures . Specifically, the scenarios represent (i) Simple heterogeneity, (ii) Complex heterogeneity, (iii) Global , and (iv) Null setting. Treatment effect heterogeneity exists in the simple and complex scenarios, whereas it is absent in the global and null scenarios. These controlled simulations allow us to systematically evaluate the performance of our method across different levels and patterns of treatment effect variation. 
Further experimental details are available in the Appendix \ref{setting_no}.

\begin{figure}[htbp]
  \centering
  \includegraphics[width=1\textwidth]{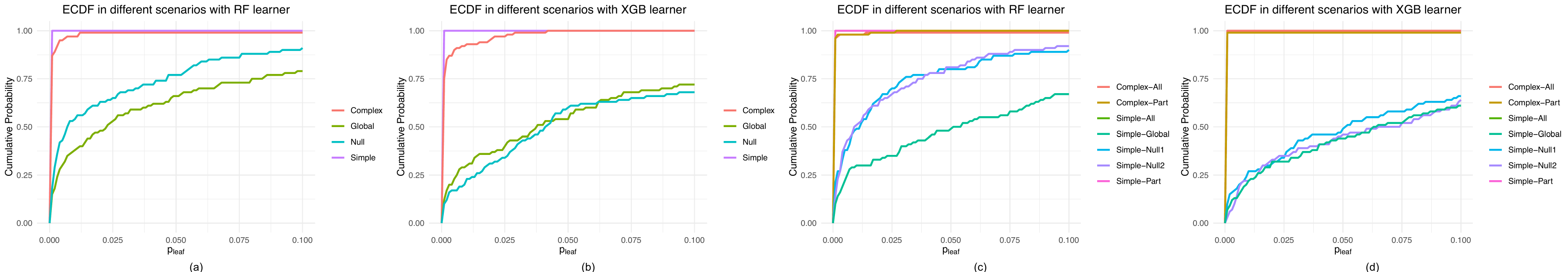}
  \caption{Empirical cumulative distribution functions (ECDF) of \(p_{leaf}\) under different scenarios, using RF and XGB as base learners. These results illustrate the sensitivity of each base learner to underlying treatment effect heterogeneity across varying levels of complexity.(a) results for RF without a mediator, (b) results for XGB without a mediator, (c) results for RF with a mediator, (d) results for XGB with a mediator.}
  \label{ecdf_nomediator}
\end{figure}

 
Table \ref{Hit_no_mediator} presents the simulation results of the unmediated model based on \(100\) experimental replications, using two different base learners, and reports the frequency with which covariates \(X_1\)and \(X_2\) are included in the identified final subtypes.
These results are obtained through a calibration-based procedure. Specifically, we determine a threshold under the Null scenario such that the Type I error rate is controlled at \(10\%\), and apply this threshold to assess the validity of subtype groupings in other scenarios. The empirical cumulative distribution functions (ECDF) of the p-values under different scenarios are illustrated in Figure \ref{ecdf_nomediator} (a) and (b), the figures elucidates why the \(p_{leaf}\) is instrumental in discerning non-heterogeneous scenarios.
 In addition, the distribution of the number of covariates contained in the final decision trees is shown in the Appendix \ref{additional_results_mlearner} Table \ref{no_mediator_dis}.
\begin{table}[h]
\caption{The table summarizes the correct covariates in profiles in each scenario. \(X_1/X_2\):Final profile contain \(X_1/X_2\), \(X_1\&X_2\): final profile contain both \(X_1\) and \(X_2\),\(X_1\) and \(X_2\) are the two variables associated with treatment effect heterogeneity. Each value denotes the count of occurrences across 100 simulations. }
    \centering
    \begin{tabular}{>{\centering\arraybackslash}p{2cm} 
                    >{\centering\arraybackslash}p{1cm}  
                    >{\centering\arraybackslash}p{1cm}
                    >{\centering\arraybackslash}p{1.5cm}
                    >{\centering\arraybackslash}p{0.01cm}
                    >{\centering\arraybackslash}p{1cm}  
                    >{\centering\arraybackslash}p{1cm}
                    >{\centering\arraybackslash}p{1.5cm}
                    }
        \hline
        Base Learner&\multicolumn{3}{c}{Random Forest}&&\multicolumn{3}{c}{XGBoost}\\
        \hline
                Covariates&\(X_1\)&\(X_2\)&\(X_1\&X_2\)&&\(X_1\)&\(X_2\)&\(X_1\&X_2\)\\
                \hline
                Simple&100&100&100&&100&100&100\\
                Complex &68&63&57&&70&70&68\\
                Global &0&2&0&&0&2&0\\
                Null &4&4&2&&1&4&1\\
                \hline
		\end{tabular}
        \label{Hit_no_mediator}
\end{table}

These results indicate that when the underlying heterogeneity is relatively simple, both RF and XGB perform well in identifying heterogeneous treatment groups.
As the heterogeneity structure increases in complexity, the performance of the proposed method deteriorates. However, as the complexity of the heterogeneity structure increases, XGB consistently outperforms RF. 
Moreover, our proposed method not only effectively identifies heterogeneous treatment groups but also maintains a low Type I error rate, underscoring its robustness and reliability across a variety of conditions. We compared our algorithm with K-Means, and the results show that K-Means lacks the ability to identify heterogeneous groups (Appendix \ref{Kmeans_algorithm} provides detailed procedures and results of the comparative algorithms can be found in Appendix \ref{Kmeans_algorithm} Table  \ref{Hit_no_mediator_kmeans},\ref{hit_kmeans_nomediator_dis} and Figure \ref{calibration_kmeans}, Figure \ref{bound} shows the boundary of the selected heterogeneous treatment region).
\subsection{Subgroup analysis with a mediator}
To identify subgroups that benefit from the treatment effects through the mediator, we design seven scenarios that reflect varying real-world heterogeneity structures . Specifically, the scenarios represent (i) simple heterogeneity, all treatment effect via mediator (Simple-All);
(ii) simple heterogeneity,part treatment effect via mediator (Simple-Part);
(iii) complex heterogeneity,all treatment effect via mediator (Complex-All);
(iv) complex heterogeneity, part treatment effect via mediator (Complex-Part);
(v) simple no heterogeneity, no treatment effect via mediator (Simple-Null1);
(vi) simple no heterogeneity, no treatment effect via mediator (Simple-Null2);
(vii) simple no heterogeneity, all units benefit from the treatment and all treatment 
effect via mediator (Simple-Global); 
 further experimental details are available in the Appendix \ref{setting_with}.
  
  Table \ref{Hit_mediator} presents the results from 100 replicated simulations of the mediation model, using two different base learners, and reports the frequency with which covariates \(X_1\) and \(X_2\) are included in the identified final subtypes. Table \ref{mediator_dis_heter} presents a comparison between the subtypes identified by the M-learner and the true heterogeneous regions, based on sample size and mediation proportion, across four heterogeneity scenarios. As shown in Table \ref{mediator_dis_heter}, the subtype regions identified by the M-learner are smaller than the true heterogeneous regions; however, the estimated mediation proportions closely approximate those of the true regions. These results suggest that the M-learner is effective in identifying subtype regions based on the indirect treatment effect (for the detailed computation methods of Table \ref{mediator_dis_heter}, please refer to the Appendix \ref{setting_with}).
  The distribution of the number of covariates contained in the final decision trees is shown in the Appendix \ref{additional_results_mlearner} Table \ref{mediator_dis}. In contrast to the unmediated setting, we calibrate the threshold based on Scenario Simple-Null2 by controlling the Type I error at \(10\%\), and then apply this threshold to assess the validity of subtype identification across other scenarios.
\begin{table}[h]
\caption{The table summarizes the correct covariates in profiles in each scenario. \(X_1/X_2\):Final profile contain \(X_1/X_2\), \(X_1\&X_2\): final profile contain both \(X_1\) and \(X_2\), \(X_1\) and \(X_2\) are the two variables associated with treatment effect heterogeneity. Each value denotes the count of occurrences across 100 simulations. }
    \centering
    \begin{tabular}{>{\centering\arraybackslash}p{2.3cm} 
                    >{\centering\arraybackslash}p{1cm}  
                    >{\centering\arraybackslash}p{1cm}
                    >{\centering\arraybackslash}p{1.5cm}
                    >{\centering\arraybackslash}p{0.01cm}
                    >{\centering\arraybackslash}p{1cm}  
                    >{\centering\arraybackslash}p{1cm}
                    >{\centering\arraybackslash}p{1.5cm}
                    }
        \hline
        &\multicolumn{3}{c}{Random Forest}&&\multicolumn{3}{c}{XGBoost}\\
        \hline
                Covariates&\(X_1\)&\(X_2\)&\(X_1\&X_2\)&&\(X_1\)&\(X_2\)&\(X_1\&X_2\)\\
                \hline
                Simple-All&99&99&98&&100&100&100\\
                Simple-Part&100&99&99&&100&100&100\\
                Complex-All&84&81&69&&98&99&97\\
                Complex-Part&78&72&54&&95&97&93\\
                Simple-Null1&9&6&4&&8&4&2\\
                Simple-Null2&4&4&1&&2&1&0\\
                Simple-Global&2&2&1&&3&5&1\\
                \hline
		\end{tabular}
        \label{Hit_mediator}
\end{table}
  \begin{table}[h]
    \centering
    \caption{ 
        Comparison of true and M-learner–estimated mediation proportions and sample sizes within heterogeneous treatment effect regions across scenarios.
        Standard deviations are shown in parentheses. 
        }
    \begin{tabular}{>{\centering\arraybackslash}p{2.0cm} 
                    >{\centering\arraybackslash}p{1.5cm}
                    >{\centering\arraybackslash}p{1.5cm}
                    >{\centering\arraybackslash}p{1.5cm}
                    >{\centering\arraybackslash}p{1.5cm}
                    >{\centering\arraybackslash}p{1.5cm}
                    >{\centering\arraybackslash}p{1.5cm}
                    }
        \hline
        Scenario&\multicolumn{2}{c}{Ground Truth}&\multicolumn{2}{c}{Random Forest}&\multicolumn{2}{c}{XGBoost}\\
        \hline
        
        &N&Med Prop&N&Med Prop&N&Med Prop\\
        \hline
        Simple-All&251(13)&1.37(0.09)&214(46)&1.33(0.10)&229(50)&1.33(0.11)\\
        Simple-Part&251(13)&0.76(0.03)&219(47)&0.78(0.06)&221(39)&0.78(0.07)\\
        Complex-All&251(13)&1.54(0.30)&221(89)&1.29(0.19)&192(64)&1.36(0.22)\\
        Complex-Part&251(13)&0.85(0.08)&216(84)&0.91(0.12)&187(95)&0.84(0.13)\\
        \hline
		\end{tabular}
        \label{mediator_dis_heter}
\end{table}

In scenarios with a simple underlying structure, both learners are capable of effectively identifying subgroups reflecting treatment effects transmitted through the mediator. However, under more complex settings, XGBoost consistently outperforms Random Forest. Specifically, the M-learner framework with XGBoost demonstrates strong capacity to detect indirect treatment effects and to recover meaningful subgroups accordingly. It also robustly rejects cases where the indirect effect is zero, indicating an ineffective mediator. Simple-Null1 and Simple-Null2 both assume a null indirect effect, but differ in whether the mediator influences the outcome. In Simple-Null1, the treatment does not affect the mediator, while in Simple-Null2, the mediator has no association with the outcome, implying it is not a true mediator. Our proposed method successfully identifies both types of scenarios using either base learner. For non-heterogeneous settings, we simulate only simple scenarios, while in heterogeneous cases, we consider both simple and complex structures to reflect realistic uncertainties regarding the complexity of heterogeneity. Simple scenarios serve as a baseline to filter out non-heterogeneous structures. Overall, the experimental results strongly support our hypotheses: across varying degrees of complexity, the proposed approach reliably determines the effectiveness of the mediator 
\(M\), estimates the mediation effect, and identifies the corresponding subgroups. Figure \ref{ecdf_nomediator} (c) and (d) show the ECDFs of p-values under different scenarios. The figure elucidates why the \(p_{leaf}\) is instrumental in discerning non-heterogeneous scenarios and in assessing the efficacy of a mediator.  Appendix Figure \ref{bound} shows the selected heterogeneous treatment regions for 100 replications.
 We compared our algorithm with K-Means, and the results show that K-Means lacks the ability to identify heterogeneous groups in the mediation model(the Appendix \ref{Kmeans_algorithm} provides detailed procedures and results of the comparative algorithms can be found in Appendix Table  \ref{Hit_mediator_kmeans_dis},\ref{kmeans_mediator_dis} ,\ref{kmeans_mediator_dis_heter} and Figure \ref{calibration_kmeans}). We further compared our method against X-learner, R-learner, and TARNet in the absence of mediators (Appendix \ref{meta_learner}), the clustering method comparsion can be found in Appendix \ref{border_clustering}.


\subsection{Sensitivity and visualization of the M-learner }
To assess the robustness of the M-learner, we systematically evaluated its sensitivity to sample size, noise level, number of clusters, projection dimension, projection technique, and clustering algorithm.

To assess the performance of the M-learner under varying sample sizes, we additionally compare the results to those obtained with a sample size of \(500\), while holding all other conditions constant. In Appendix \ref{additional_results_mlearner}, the Table \ref{Hit_no_mediator_500}-\ref{mediator_dis_heter_500} and the Figure \ref{ecdf_sample_size}-\ref{xgb_71_p_mediator} show the detailed results. The results indicate that, in the absence of mediators, the M-learner is capable of accurately identifying heterogeneous treatment regions in the simple scenario, even with a reduced sample size. However, its performance deteriorates substantially in the complex scenario, suggesting that the identification of treatment effect heterogeneity under more complex conditions requires larger sample sizes. Notably, the M-learner continues to perform well in detecting non-heterogeneous scenarios, even when the available sample size is limited. 

In the presence of a mediator, the M-learner remains effective in identifying heterogeneous treatment regions mediated by the mediator, even with a reduced sample size. In the complex scenario, the performance of the RF-based learner declines significantly, whereas the performance degradation of the XGB-based learner is relatively limited. This suggests that, under limited sample sizes, XGB serves as a more robust base learner. The M-learner also demonstrates stable performance; however, accurately estimating complex relationships among \(M\), \(X\), and \(Y\) still requires a sufficiently large sample size to reliably identify heterogeneous treatment effects.

Comparisons of  ECDFs across varying sample sizes and scenarios reveal that, in the simple scenario, the ECDFs for sample sizes of \(1000\) and \(500\) exhibit minimal divergence. In contrast, more pronounced differences are observed in the complex scenario. Notably, an ECDF curve positioned closer to the upper-left corner reflects stronger treatment effect heterogeneity, whereas a curve nearer to the lower-right corner indicates weaker heterogeneity. A comprehensive comparison across different learners, scenarios, and sample sizes is presented in Appendix Figure \ref{rf_41_p_no_mediator}-\ref{xgb_71_p_mediator}.

The visualization results of the M-learner are provided in Appendix \ref{Visualization} for reference. Further sensitivity analysis experiments can be found in the Appendix \ref{sensitivity}.

\section{Real Data Application}
In this section, we apply the proposed method to analyze the JOBS II real dataset, collected from a randomized of a job training intervention on unemployed workers. The dataset can be downloaded from R package "mediation"\citep{tingley2014mediation}. The JOBS II study is a well-known randomized controlled trial conducted in the United States, designed to evaluate the effectiveness of a job search intervention program for unemployed individuals. The study enrolled 899 participants, who were randomly assigned either to a job training intervention group or to a control group. The dataset includes a rich set of covariates, such as demographic information and psychological measures (e.g., self-efficacy, depression). In follow-up interviews, the outcome—a continuous measure of depressive symptoms was assessed. The mediator, \(M\), is a continuous measure representing job search self-efficacy\citep{price1992impact,vinokur2000two,vuori2005benefits,cheng2022causal}.

We use age, sex, education, prior occupational status, and the level of economic hardship experienced by participants as covariates. For educational background, we classify participants into three categories: individuals who did not complete high school or whose highest degree is a high school diploma are grouped together; individuals who attended some college but did not obtain a bachelor's degree form the second group; and those who earned a bachelor's degree or higher are classified into the third group. Other covariates are kept in their original form. 

We first conducted a preliminary analysis of the JOBS II data using the  mediation package\citep{imai2010general}. Table \ref{real_data_table}
shows the detailed analysis. Job search self-efficacy has been widely employed as a mediator in the literature\citep{vinokur1995impact,vinokur2000two}. However, it exhibits relatively large p-value. 

\begin{figure}[h]
  \centering
    \includegraphics[width=1\linewidth]{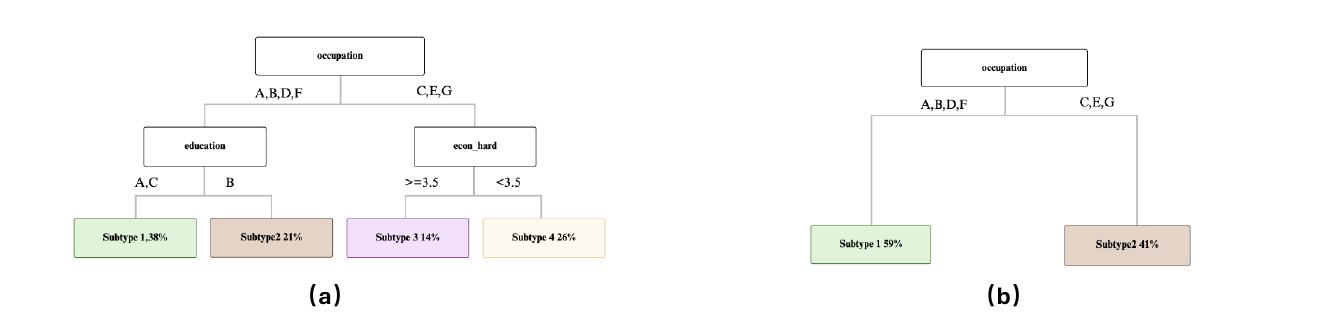}
  \caption{Subtype results identified by the M-learner on the Jobs II dataset. (a)
  It shows subtype results obtained without mediators, (b) it shows subtype results obtained with the mediator. In each decision tree, each leaf represents a subtype. econ\_hard refers to the level of economic hardship. In occupation, A denotes "clerical\/kindred",B denotes "laborers\/service works",C denotes "operatives\/kindred works",D denotes"sales workers",E denotes "craftsmen\/foremen\/kindred",F denotes "manegerial", G denotes "professionals".In education, A denotes individuals who did not complete high school or whose highest degree is a high school
diploma, B denotes individuals who attended some college but did not obtain a bachelor’s
degree, C denotes those who earned a bachelor’s degree or higher.}
  \label{real_data}
\end{figure}

Applying our proposed method to the JOBS II data, we first analyzed the heterogeneity based solely on the TTE. We then applied the mediation analysis separately to each subtype to estimate the TTE, ITE, mediation proportions, and corresponding p-values.  Figure \ref{real_data}, specifically the first decision tree, identifies four subtypes. Subtype 1 and 2 are associated with occupation and the level of economic hardship. Subtype 1 includes individuals from occupation categories A, B, D, and F, as well as those with either higher or lower levels of education. Subtype 2 consists of individuals from the same occupation categories (A, B, D, F) but with moderate levels of education. According to the results in Table \ref{real_data_table}, subtype 1 shows significant TTE, ITE, and mediation proportions. In this subtype, receiving job search assistance (treatment) significantly improves the mood, with \(M\) identified as a significant mediator. In contrast, subtype 2 exhibits a weaker treatment effect, suggesting that job search assistance may have a limited impact on this subtype, with \(M\) showing no significant mediating role.
Subtype 3 and 4 are also associated with occupation and economic hardship. Subtype 3 is characterized by individuals in occupation categories C, E, and G, as well as those experiencing greater economic difficulties, while subtype 4 includes individuals from the same occupation categories (C, E, G) but with relatively better economic conditions. In Subtype 3, both TTE and ITE are observed, though the p-values are relatively large, which may be attributable to the smaller sample size. For subtype 4, the TTE is significant, but the ITE is not, along with a lower mediation proportion. Interestingly, when not receiving job search assistance, the mood of unemployed individuals in this group tends to improve.

Employing a mediation analysis approach of M-learner, as depicted in the second decision tree of Figure \ref{real_data}, yielded two distinct subtypes. In contrast to findings derived without considering a mediator, these subtypes exhibited a broader scope and were exclusively associated with occupational factors. Specifically, subtype 1 in this analysis represents the aggregation of subtypes 1 and 2 from the initial, non-mediated analysis, while subtype 2 comprises the combined subtypes 3 and 4. As presented in Table \ref{real_data_table}, subtype 1 demonstrated significant TTE and ITE, accompanied by a substantial proportion of mediation and a reduced p-value. Conversely, subtype 2 exhibited smaller and statistically non-significant TTE and ITE.
\begin{table}[H]
    \centering
    \caption{ 
        Summary of the total treatment effect, indirect treatment effect, mediation proportion, and corresponding p-values, calculated using methods from the "mediation" package \cite{imai2010general}and the proposed approach. Med prop denotes mediation proportion\(:=\)indirect treatment effect\(/\) total treatment effect. TTE denotes total treatment effect
        The value in parentheses is the p-value. ITE denotes indirect treatment effect.
        The value in parentheses is the p-value. N denotes sample size.
        }
    \begin{tabular}{>{\centering\arraybackslash}p{2cm} 
                    >{\centering\arraybackslash}p{1.3cm} 
                    >{\centering\arraybackslash}p{0.6cm}
                    >{\centering\arraybackslash}p{2.2cm}
                    >{\centering\arraybackslash}p{2.2cm}
                    >{\centering\arraybackslash}p{2cm}
                    }
        \hline
        Method&Subtype&N&TTE&ITE&Med Prop\\
        \hline
        "Mediation"&NA&\(899\)&\(-0.063(0.29)\)&\(-0.015(0.21)\)&\(24.0\%(0.29)\)\\
        \hline
        \multirow{4}{*}{\makecell{M-learner,\\ no mediator}}&Subtype1&\(344\)&\(-0.243(0.004)\)&\(-0.057(0.010)\)&\(23.6\%(0.014)\)\\
                                            &Subtype2&\(190\)&\(0.003(0.95\))&\(0.005(0.77)\)&\(160\%(0.81)\)\\
                                            &Subtype3&\(129\)&\(-0.10(0.42\))&\(-0.013(0.76)\)&\(12.9\%(0.73)\)\\
                                            &Subtype4&\(236\)&\(0.145(0.03\))&\(0.009(0.47)\)&\(6.1\%(0.48)\)\\                 \hline  
        \multirow{2}{*}{\makecell{M-learner,\\ with mediator}}&Subtype1&\(534\)&\(-0.15(0.016)\)&\(-0.030(0.010)\)&\(20.0\%(0.052)\)\\
                                            &Subtype2&\(365\)&\(0.065(0.30\))&\(0.005(0.75)\)&\(8.0\%(0.74)\)\\
        \hline
		\end{tabular}
        \label{real_data_table}
\end{table}
These results suggest that the mediator exhibits heterogeneity across different subtypes and serves as an effective mediator only in certain groups. Regardless of whether the mediator is included, our method yielded similar subgroup structures, with the key difference being the further subdivision of certain occupational groups when considering the mediator. Using this approach, in future government-initiated randomized clinical trials, it would be possible to monitor changes in the mediator within specific subtypes to quickly assess whether a new intervention is effective for those groups. Trials could be stopped early for ineffective interventions in particular groups, allowing better, more targeted treatments to be administered. This strategy can significantly reduce government costs while enabling timely adjustments to intervention strategies, thereby minimizing the negative impact of ineffective treatments on participants.

\section{Discussion}
This article introduced a novel method
that is, to our knowledge, the first to examine treatment effect heterogeneity transmitted through the mediator.  In addition to estimating mediated heterogeneity, our approach enables data-driven subgroup identification based on distinct mediator patterns. This represents a significant methodological advancement, as it allows researchers to evaluate the informativeness of mediators and to detect heterogeneous treatment effects that are revealed through them. The proposed framework is highly flexible and can accommodate a variety of base learners, including random forests, XGBoost, and neural networks, depending on the data characteristics and specific application context. Moreover, the method is broadly applicable to both mediated and unmediated settings. In the absence of a mediator, it enables estimation of the CATTE and facilitates the detection of heterogeneity and subgroup structures driven by the TTE. When a mediator is present, the method estimates the CAITE and identifies heterogeneity and subgroup structures that arise through the mediator. By capturing treatment effect heterogeneity from both total and mediated perspectives, the framework offers a comprehensive understanding of the complex interplay among treatment, covariates, mediators, and outcomes.

Our method is unbiased when the \(Y\) has linear relationship with mediator \(M\), for nonlinearship, it has bias. However, with faster convergence learners, the bias of the estimator can diminish as the sample size increases.

Importantly, our method permits the inference of individual treatment responsiveness without requiring observation of the final outcome variable \(Y\). This finding holds substantial implications across disciplines such as economics, psychology, medicine and sociology. In practical applications, such as in technology companies, mediator behavior can guide personalized interventions. In healthcare, particularly in pharmaceutical settings, the method can assist clinicians in adapting treatment strategies, thereby contributing to the development of precision medicine.

A potential direction for future research is to extend our method to survival analysis settings, which could enable the pharmaceutical industry to predefine subgroups based on mediating variables, ultimately enhancing the success rate of drug development.



\bibliographystyle{plain}
\bibliography{references}


\newpage
\appendix
\section{Technical Appendices and Supplementary Material}

\subsection{Assumption}
Here, we introduce the assumption of the M-learner framework. 

For random variables $A,B$ and $C$, let $A\perp\!\!\!\perp B|C$
denote that $A$ is conditionally independent of $B$, given $C$.
\begin{eqnarray}
\label{eq1}
Y(w,m)\perp\!\!\!\perp W|X
\end{eqnarray}
for all  $w$ and $m$.

\begin{eqnarray}
\label{eq2}
Y(w,m)\perp\!\!\!\perp M|X,W
\end{eqnarray}
for all  $w$ and $m$.

\begin{eqnarray}
\label{eq3}
M(w)\perp\!\!\!\perp W|X
\end{eqnarray}
for all  $w$.

\begin{eqnarray}
\label{eq4}
Y(w,m)\perp\!\!\!\perp M(w*)|X
\end{eqnarray}
for all  $w$,$w*$,$m$.

Additionally, in the M-learner method, it assumes a randomized controlled trial (RCT) setting, the treatment assignment 
$W$ is independent of both the covariates 
$X$ and the mediator 
$M$, i.e. 
\begin{eqnarray}
\label{eq5}
   W\perp\!\!\!\perp X 
\end{eqnarray}

When there is no mediator, we can omit the \(M\) in the assumptions (\ref{eq1}) - (\ref{eq4}).

\subsection{Setting in no mediator model}
\label{setting_no}
\renewcommand{\thetable}{A.\arabic{table}}
\renewcommand{\thefigure}{A.\arabic{figure}}
Across all experimental designs in this section, we fix the sample size at \(1,000\), with \(10\) covariates generated for each unit. Subjects were randomly assigned to treatment and control groups in a \(1:1\) ratio. We specify two underlying functions: \(\eta(x)\)
, representing the conditional mean outcome, and 
\(\kappa(x)\), capturing the conditional treatment effect. Both functions are defined for units under treatment (\(w=1\))
 and control (\(w=0\)).
 For simple scenario,
 \begin{eqnarray}
 \label{simple}
     Y_i(w) = \eta(X_i)+\frac{1}{2}(2w-1)\cdot\kappa(X_i)+b+\epsilon_i,
 \end{eqnarray}
 for complex scenario,
 \begin{eqnarray}
 \label{complex}
     Y_i(w) = \frac{1}{1+\exp({\eta(X_i)+\frac{1}{2}(2w-1)\cdot\kappa(X_i)+b+\epsilon_i})},
 \end{eqnarray}
 where \(\epsilon_i\sim \mathcal{N}(0,0.01)\), and the \(X_i\) are independent of \(\epsilon_i\) and one another, and \(X_i\sim \mathcal{N}(0,1)\), \(b\) is the intercept.  The four scenarios designs follow:
 \begin{itemize}
 \item[1.] Simple scenario in (\ref{simple}) , existing heterogeneity:\(\eta(x) = \frac{1}{2}(x_1+x_2)+x_3+x_4\),\(\kappa(x)=\sum_{i=1}^{2}\mathbb{I}(x_i>0)\cdot x_i\), \(b=1\).
  \item[2.] Complex scenario in (\ref{complex}), existing heterogeneity:\(\eta(x) = \frac{1}{2}(x_1+x_2)+x_3+x_4\),\(\kappa(x)=\sum_{i=1}^{2}\mathbb{I}(x_i>0)\cdot x_i\), \(b=1\).
    \item[3.] Simple scenario in (\ref{simple}), no heterogeneity,all units benefit from treatment(Global):\(\eta(x) = x_3+x_4\),\(\kappa(x)=1\), \(b=1\).
      \item[4.] Simple scenario in (\ref{simple}), no heterogeneity, no units benefit from treatment(NULL):\(\eta(x) = x_3+x_4\),\(\kappa(x)=0\), \(b=1\).
 \end{itemize}
For the RF model, the number of trees is set to 2000, with all other parameters kept at their default values. For XGBoost, the number of boosting rounds is set to 100, while all remaining parameters were left at their default settings. In all simulation experiments, the range of cluster numbers was predefined as \(2\) to \(5\). 

In this paper, all experiments were conducted on a MacBook Pro equipped with an M3 Max CPU and 36GB of RAM. The software environment includes R version 4.3.3, with the randomForestSRC package version 3.2.3, xgboost version 1.7.9.1 and rpart version 4.1.23.

\subsection{Setting in mediator model}
\label{setting_with}
Across all experimental designs in mediator model, we fixed the sample size at \(1,000\), with \(10\) covariates generated for each unit. Subjects were randomly assigned to treatment and control groups in a \(1:1\) ratio. We specified four underlying functions: \(\eta_1(x)\)
, representing the conditional mean outcome in mediator covariates model, \(\eta_2(x)\)
, representing the conditional mean outcome in covariates reponse model, and \(\kappa_1(x)\), capturing the conditional treatment effect in mediator covariates model, and 
\(\kappa_2(x)\), capturing the conditional treatment effect in  covariates reponse model.
 All functions were defined for units under treatment (\(w=1\))
 and control (\(w=0\)), 
 for simple scenario,
 \begin{eqnarray}
 \label{simple_mediator}
 M_i(w) &=&\eta_1(X_i)+\frac{1}{2}(2w-1)\cdot\kappa_1(X_i)+b_1+\epsilon_{1i},\nonumber\\
     Y_i(w) &=& \eta_2(X_i)+\frac{1}{2}(2w-1)\cdot\kappa_2(X_i)+b_2+c\cdot M_i(w)+\epsilon_{2i}.
 \end{eqnarray}
 For complex scenario,
 \begin{eqnarray}
 \label{complex_mediator}
  M_i(w) &=&\eta_1(X_i)+\frac{1}{2}(2w-1)\cdot\kappa_1(X_i)+b_1+\epsilon_{1i},\nonumber\\
     Y_i(w) &=& \frac{1}{1+\exp({\eta_2(X_i)+\frac{1}{2}(2w-1)\cdot\kappa_2(X_i)+b_2+c\cdot M_i(w)+\epsilon_{2i})}},
 \end{eqnarray}
 where \(\epsilon_{1i}\sim \mathcal{N}(0,0.01)\), \(\epsilon_{2i}\sim \mathcal{N}(0,0.01)\), and the \(X_i\) are independent of \(\epsilon_{1i}\), \(\epsilon_{2i}\) and one another, and \(X_i\sim \mathcal{N}(0,1)\), \(b_1\) and \(b_2\) are the intercept terms, \(c\) is the coefficients of mediator.
The seven scenarios designs follow:
 \begin{itemize}
 \item[1.] Simple scenario in (\ref{simple_mediator}),
 existing heterogeneity,all treatment effects via mediator:\(\eta_1(x) = \frac{1}{2}(x_1+x_2)+x_3+x_4\),\(\kappa_1(x)=\sum_{i=1}^{2}\mathbb{I}(x_i>0)\cdot x_i\), \(b_1=0\),
 \(\eta_2(x) = \frac{1}{2}(x_3+x_4)\),\(\kappa_2(x)=0\), \(b_2=1, c = 1\).
  \item[2.] Simple scenario in (\ref{simple_mediator}), existing heterogeneity, part of the treatment effect via mediator:
  \(\eta_1(x) = \frac{1}{2}(x_1+x_2)+x_3+x_4\),\(\kappa_1(x)=\sum_{i=1}^{2}\mathbb{I}(x_i>0)\cdot x_i\), \(b_1=0, c = 1\),
 \(\eta_2(x) = \frac{1}{2}(x_3+x_4)\),\(\kappa_2(x)=\sum_{i=1}^{2}\mathbb{I}(x_i>0)\cdot x_i\), \(b_2=1, c = 1\).
  \item[3.]Complex scenario in (\ref{complex_mediator}), existing heterogeneity, all treatment effects via mediator:\(\eta_1(x) = \frac{1}{2}(x_1+x_2)+x_3+x_4\),\(\kappa_1(x)=\sum_{i=1}^{2}\mathbb{I}(x_i>0)\cdot x_i\), \(b_1=0\),
 \(\eta_2(x) = \frac{1}{2}(x_3+x_4)\),\(\kappa_2(x)=0\), \(b_2=1, c = 1\).
  \item[4.] Complex scenario in (\ref{complex_mediator}), existing heterogeneity, part of the treatment effect via mediator: \(\eta_1(x) = \frac{1}{2}(x_1+x_2)+x_3+x_4\),\(\kappa_1(x)=\sum_{i=1}^{2}\mathbb{I}(x_i>0)\cdot x_i\), \(b_1=0, c = 1\),
 \(\eta_2(x) = \frac{1}{2}(x_3+x_4)\),\(\kappa_2(x)=\sum_{i=1}^{2}\mathbb{I}(x_i>0)\cdot x_i\), \(b_2=1, c = 1\).
\item[5.] Simple scenario in (\ref{simple_mediator}), no heterogeneity, \(0\%\) treatment effects via mediator(NULL 1):\(\eta_1(x) = \frac{1}{2}(x_1+x_2)+x_3+x_4\),\(\kappa_1(x)=0\), \(b_1=0\),
 \(\eta_2(x) = \frac{1}{2}(x_3+x_4)\),\(\kappa_2(x)=\sum_{i=1}^{2}\mathbb{I}(x_i>0)\cdot x_i\), \(b_2=1,c = 1\).
 \item[6.] Simple scenario in (\ref{simple_mediator}), no heterogeneity, \(M\) is not mediator, all treatment effects are directly transmitted to \(Y\)(NULL 2):\(\eta_1(x) = \frac{1}{2}(x_1+x_2)+x_3+x_4\),\(\kappa_1(x)=0\), \(b_1=0\),
 \(\eta_2(x) = \frac{1}{2}(x_3+x_4)\),\(\kappa_2(x)=\sum_{i=1}^{2}\mathbb{I}(x_i>0)\cdot x_i\), \(b_2=1, c = 0\).
  \item[7.] Simple scenario in (\ref{simple_mediator}), no heterogeneity, all units benefit from the treatment and all treatment effects via mediator (Global):\(\eta_1(x) = \frac{1}{2}(x_1+x_2)+x_3+x_4\),\(\kappa_1(x)=1\), \(b_1=0\),
 \(\eta_2(x) = \frac{1}{2}(x_3+x_4)\),\(\kappa_2(x)=0\), \(b_2=1, c = 1\).

 \end{itemize}
 For the RF model, the number of trees is set to 2000, with all other parameters kept at their default values. For XGBoost, the number of boosting rounds is set to 100, while all remaining parameters are left at their default settings. In all simulation experiments, the range of cluster numbers is predefined as \(2\) to \(5\). 

 In the analysis presented in Table \ref{mediator_dis_heter}, the ground truth heterogeneous region is defined by samples satisfying the conditions \(X_1 > 0\) and \(X_2 > 0\). The mediation proportion for this region is calculated using the R package "mediation". To evaluate the mediation proportion and sample size corresponding to the heterogeneous region identified by the M-learner, we adopt the following procedure: among the final subtype regions generated by the decision tree, we assess the mediation effect within each region and select the region exhibiting the most statistically significant mediation effect. The sample size and mediation effect of this selected region are then recorded.

 \subsection{Additional simulation results for M-learner}
\label{additional_results_mlearner}
\begin{table}[H]
    \centering
    \caption{ 
        The table summarizes the distribution of the number of variables included in the selected decision trees across 100 simulation runs. Each value represents the frequency with which a specific number of variables was selected over the course of the experiments. The performance of Random Forest and XGBoost as base learners was systematically evaluated under four distinct experimental scenarios. Selecting zero covariates in a given experimental run indicates that the decision tree’s classification was not supported by the calibration process and was therefore rejected.
        }
    \begin{tabular}{>{\centering\arraybackslash}p{3.3cm} 
                    >{\centering\arraybackslash}p{0.4cm}  
                    >{\centering\arraybackslash}p{0.4cm}
                    >{\centering\arraybackslash}p{0.4cm}
                    >{\centering\arraybackslash}p{0.4cm}
                    >{\centering\arraybackslash}p{0.4cm}
                    >{\centering\arraybackslash}p{0.01cm}
                    >{\centering\arraybackslash}p{0.4cm}
                    >{\centering\arraybackslash}p{0.4cm}
                    >{\centering\arraybackslash}p{0.4cm}
                    >{\centering\arraybackslash}p{0.4cm}
                    >{\centering\arraybackslash}p{0.4cm}
                    }
        \hline
        Base Learner&\multicolumn{5}{c}{Random Forest}&&\multicolumn{5}{c}{XGBoost}\\
        \hline
        Number of covariates&0&1&2&3&4&&0&1&2&3&4\\
        \hline
        Simple&0&0&29&52&19&&0&0&57&28&15\\
        Complex&25&1&9&34&31&&28&1&7&30&34\\
        Global&92&0&5&2&1&&98&0&2&0&0\\
        Null&90&1&3&5&1&&90&3&3&3&1\\
        \hline  
		\end{tabular}
        \label{no_mediator_dis}
\end{table}

\begin{table}[H]
    \centering
    \caption{ 
        The table summarizes the distribution of the number of variables included in the selected decision trees across 100 simulation runs. Each value represents the frequency with which a specific number of variables was selected over the course of the experiments. The performance of Random Forest and XGBoost as base learners was systematically evaluated under seven distinct experimental scenarios. Selecting zero covariates in a given experimental run indicates that the decision tree’s classification was not supported by the calibration process and was therefore rejected.
        }
    \begin{tabular}{>{\centering\arraybackslash}p{3.0cm} 
                    >{\centering\arraybackslash}p{0.4cm}  
                    >{\centering\arraybackslash}p{0.4cm}
                    >{\centering\arraybackslash}p{0.4cm}
                    >{\centering\arraybackslash}p{0.4cm}
                    >{\centering\arraybackslash}p{0.4cm}
                    >{\centering\arraybackslash}p{0.4cm}
                    >{\centering\arraybackslash}p{0.01cm}
                    >{\centering\arraybackslash}p{0.4cm}
                    >{\centering\arraybackslash}p{0.4cm}
                    >{\centering\arraybackslash}p{0.4cm}
                    >{\centering\arraybackslash}p{0.4cm}
                    >{\centering\arraybackslash}p{0.4cm}
                    >{\centering\arraybackslash}p{0.4cm}
                    }
        \hline
        Base Learner&\multicolumn{6}{c}{Random Forest}&&\multicolumn{6}{c}{XGBoost}\\
        \hline
        Number of covariates&0&1&2&3&4&\(\geq\)5&&0&1&2&3&4&\(\geq\)5\\
        \hline
         Simple-All&0&0&35&47&17&1&&0&0&58&33&9&0\\
         Simple-Part&0&0&34&51&14&1&&0&0&54&40&6&0\\
         Complex-All&4&1&10&36&48&1&&0&1&23&43&33&0\\
         Complex-Part&4&1&8&51&36&0&&1&1&14&49&35&0\\
         Simple-Null1&84&1&6&7&2&0&&83&1&8&6&1&1\\
         Simple-Null2&90&1&4&4&1&0&&90&5&3&2&0&0\\
         Simple-Global&94&0&2&3&1&0&&87&1&6&4&2&0\\
        \hline  
		\end{tabular}
        \label{mediator_dis}
\end{table}

\begin{table}[H]
\caption{The table summarizes the correct covariates in profiles in each scenario when sample size is \(500\) in no mediators model. \(X_1/X_2\):Final profile contain \(X_1/X_2\), \(X_1\&X_2\): final profile contain both \(X_1\) and \(X_2\),\(X_1\) and \(X_2\) are the two variables associated with treatment effect heterogeneity. Each value denotes the count of occurrences across 100 simulations. }
    \centering
        \begin{tabular}{>{\centering\arraybackslash}p{2cm} 
                    >{\centering\arraybackslash}p{1cm}  
                    >{\centering\arraybackslash}p{1cm}
                    >{\centering\arraybackslash}p{1.5cm}
                    >{\centering\arraybackslash}p{0.01cm}
                    >{\centering\arraybackslash}p{1cm}  
                    >{\centering\arraybackslash}p{1cm}
                    >{\centering\arraybackslash}p{1.5cm}
                    }
        \hline
        Base Learner&\multicolumn{3}{c}{Random Forest}&&\multicolumn{3}{c}{XGBoost}\\
        \hline
                Covariates&\(X_1\)&\(X_2\)&\(X_1\&X_2\)&&\(X_1\)&\(X_2\)&\(X_1\&X_2\)\\
                \hline
                Simple&95&94&93&&96&96&95\\
                Complex &26&24&19&&33&25&19\\
                Global &4&3&1&&0&0&0\\
                Null &5&4&2&&4&3&1\\
                \hline
		\end{tabular}
                \label{Hit_no_mediator_500}
        
\end{table}

\begin{table}[H]
    \centering
    \caption{ 
        The table summarizes the distribution of the number of variables included in the selected decision trees across 100 simulation runs when there is no mediator and sample size is \(500\). Each value represents the frequency with which a specific number of variables was selected over the course of the experiments. The performance of Random Forest and XGBoost as base learners was systematically evaluated under four distinct experimental scenarios. Selecting zero covariates in a given experimental run indicates that the decision tree’s classification was not supported by the calibration process and was therefore rejected.
        }
    \begin{tabular}{>{\centering\arraybackslash}p{3.0cm} 
                    >{\centering\arraybackslash}p{0.4cm}  
                    >{\centering\arraybackslash}p{0.4cm}
                    >{\centering\arraybackslash}p{0.4cm}
                    >{\centering\arraybackslash}p{0.4cm}
                    >{\centering\arraybackslash}p{0.6cm}
                    >{\centering\arraybackslash}p{0.001cm}
                    >{\centering\arraybackslash}p{0.4cm}
                    >{\centering\arraybackslash}p{0.4cm}
                    >{\centering\arraybackslash}p{0.4cm}
                    >{\centering\arraybackslash}p{0.4cm}
                    >{\centering\arraybackslash}p{0.6cm}
                    }
        \hline
        Base Learner&\multicolumn{5}{c}{Random Forest}&&\multicolumn{5}{c}{XGBoost}\\
        \hline
        Number of covariates&0&1&2&3&\(\geq\)4&&0&1&2&3&\(\geq\)4\\
        \hline
        Simple&4&1&42&37&16&&3&1&70&21&5\\
        Complex&67&3&9&11&10&&61&3&7&18&11\\
        Global&91&0&3&5&2&&99&0&1&0&0\\
        Null&90&2&2&4&5&&90&2&3&3&2\\
        \hline  
		\end{tabular}
        \label{no_mediator_dis_500}
\end{table}


\begin{table}[H]
\caption{The table summarizes the correct covariates in profiles in each scenario when sample size is \(500\). \(X_1/X_2\):Final profile contain \(X_1/X_2\), \(X_1\&X_2\): final profile contain both \(X_1\) and \(X_2\), \(X_1\) and \(X_2\) are the two variables associated with treatment effect heterogeneity. Each value denotes the count of occurrences across 100 simulations. }
    \centering
    \begin{tabular}{>{\centering\arraybackslash}p{2.3cm} 
                    >{\centering\arraybackslash}p{1cm}  
                    >{\centering\arraybackslash}p{1cm}
                    >{\centering\arraybackslash}p{1.5cm}
                    >{\centering\arraybackslash}p{0.01cm}
                    >{\centering\arraybackslash}p{1cm}  
                    >{\centering\arraybackslash}p{1cm}
                    >{\centering\arraybackslash}p{1.5cm}
                    }
        \hline
        Base Learner&\multicolumn{3}{c}{Random Forest}&&\multicolumn{3}{c}{XGBoost}\\
        \hline
                Covariates&\(X_1\)&\(X_2\)&\(X_1\&X_2\)&&\(X_1\)&\(X_2\)&\(X_1\&X_2\)\\
                \hline
                Simple-All&93&90&89&&92&94&91\\
                Simple-Part&97&96&96&&92&93&92\\
                Complex-All&46&39&23&&86&83&76\\
                Complex-Part&38&33&13&&76&69&57\\
                Simple-Null1&9&5&3&&0&1&0\\
                Simple-Null2&6&2&1&&2&2&1\\
                Simple-Global&2&0&0&&0&2&0\\
                \hline
		\end{tabular}
        \label{Hit_mediator_500}
\end{table}

\begin{table}[H]
    \centering
    \caption{ 
        The table summarizes the distribution of the number of variables included in the selected decision trees across 100 simulation runs when the sample size is \(500\). Each value represents the frequency with which a specific number of variables was selected over the course of the experiments. The performance of Random Forest and XGBoost as base learners was systematically evaluated under seven distinct experimental scenarios. Selecting zero covariates in a given experimental run indicates that the decision tree’s classification was not supported by the calibration process and was therefore rejected.
        }
    \begin{tabular}{>{\centering\arraybackslash}p{3.3cm} 
                    >{\centering\arraybackslash}p{0.4cm}  
                    >{\centering\arraybackslash}p{0.4cm}
                    >{\centering\arraybackslash}p{0.4cm}
                    >{\centering\arraybackslash}p{0.4cm}
                    >{\centering\arraybackslash}p{0.4cm}
                    >{\centering\arraybackslash}p{0.4cm}
                    >{\centering\arraybackslash}p{0.01cm}
                    >{\centering\arraybackslash}p{0.4cm}
                    >{\centering\arraybackslash}p{0.4cm}
                    >{\centering\arraybackslash}p{0.4cm}
                    >{\centering\arraybackslash}p{0.4cm}
                    >{\centering\arraybackslash}p{0.4cm}
                    >{\centering\arraybackslash}p{0.4cm}
                    }
        \hline
        Base Learner&\multicolumn{6}{c}{Random Forest}&&\multicolumn{6}{c}{XGBoost}\\
        \hline
        Number of covariates&0&1&2&3&4&\(\geq\)5&&0&1&2&3&4&\(\geq\)5\\
        \hline
         Simple-All&6&3&42&35&14&0&&5&4&65&23&3&0\\
         Simple-Part&3&1&46&39&11&0&&7&1&67&18&7&0\\
         Complex-All&37&6&16&23&13&0&&7&10&25&48&9&1\\
         Complex-Part&42&5&12&29&12&0&&12&12&30&31&15&0\\
         Simple-Null1&86&2&6&3&3&0&&96&1&3&0&0&0\\
         Simple-Null2&90&2&6&2&0&0&&96&1&1&2&0&0\\
         Simple-Global&94&0&1&4&1&0&&97&0&0&2&1&0\\
        \hline  
		\end{tabular}
        \label{mediator_dis_500A}
\end{table}


\begin{table}[H]
    \centering
    \caption{ 
        Comparison of true and M-learner–estimated mediation proportions and sample sizes within heterogeneous treatment effect regions across scenarios when sample size is \(500\).Standard deviations are shown in parentheses. The mediation proportion is calculated by R package "mediation". "For each set of results, the subtype exhibiting the lowest p-value for the mediation proportion is designated as the region of interest. Within this region, we estimate the mediation proportion, total treatment effect, and indirect treatment effect.
        }
    \begin{tabular}{>{\centering\arraybackslash}p{2.0cm} 
                    >{\centering\arraybackslash}p{1.5cm}
                    >{\centering\arraybackslash}p{1.5cm}
                    >{\centering\arraybackslash}p{1.5cm}
                    >{\centering\arraybackslash}p{1.5cm}
                    >{\centering\arraybackslash}p{1.5cm}
                    >{\centering\arraybackslash}p{1.5cm}
                    }
        \hline
        Scenario&\multicolumn{2}{c}{True}&\multicolumn{2}{c}{Random Forest}&\multicolumn{2}{c}{XGBoost}\\
        \hline
        
        &N&Med Prop&N&Med Prop&N&Med Prop\\
        \hline
        Simple-All&124(10)&1.38(0.14)&111(28)&1.27(0.13)&114(30)&1.29(0.15)\\
        Simple-Part&124(10)&0.76(0.04)&113(25)&0.78(0.06)&113(23)&0.77(0.04)\\
        Complex-All&124(10)&1.64(0.69)&140(66)&1.29(0.19)&116(55)&1.25(0.23)\\
        Complex-Part&124(10)&0.87(0.12)&134(59)&0.91(0.12)&136(61)&0.93(0.18)\\
        \hline
		\end{tabular}
        \label{mediator_dis_heter_500}
\end{table}

\begin{figure}[H]
  \centering
  \includegraphics[width=1\textwidth]{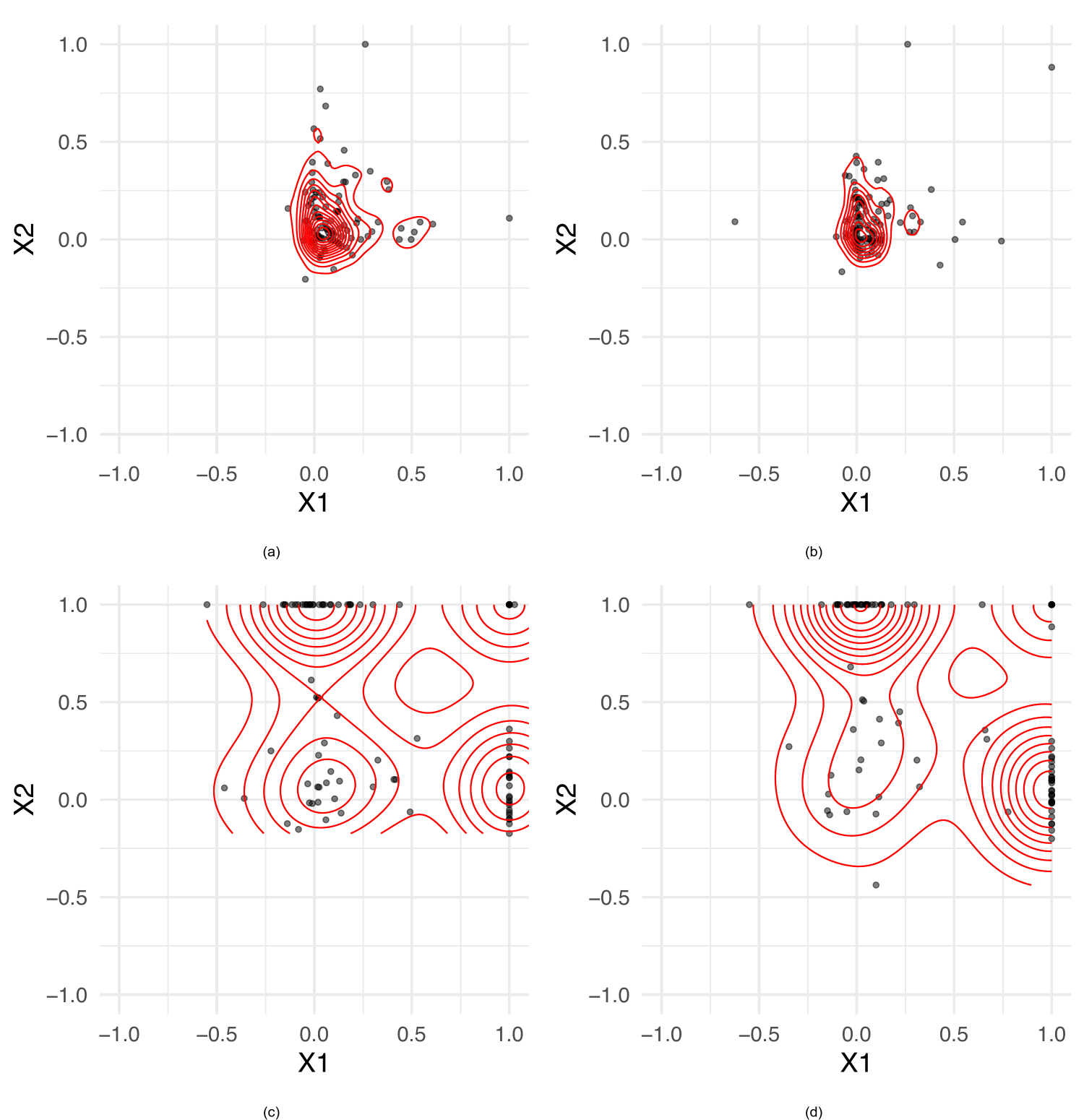}
  \caption{Threshold location distribution plot for Simple-All scenario. dot in the figure represents the threshold of heterogeneous region for 100 replications, red line represents the density line. If a variable was not selected or the threshold exceeded 1, we assigned a value of 1. (a) Simple-All, (b) Simple-Part, (c) Complex-All, (d) Complex-Part.}
  \label{bound}
\end{figure}

 \begin{figure}[H]
  \centering
  \includegraphics[width=0.88\textwidth]{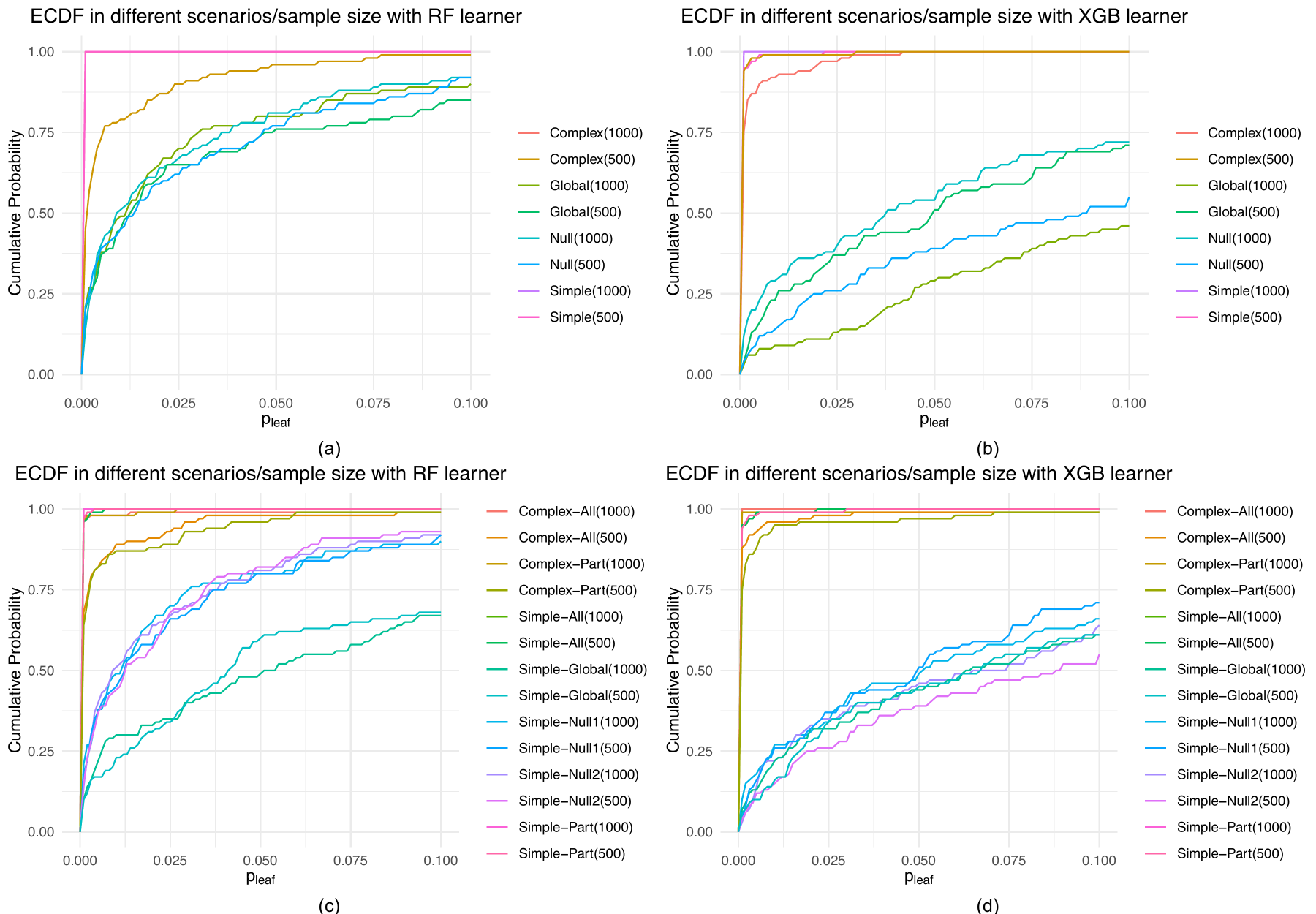}
  \caption{Comparison of \(p_{leaf}\) ECDFs using RF and XGB as base learners across different scenarios with sample sizes of \(1000\) and \(500\).(a)results for RF without a mediator, (b) results for XGB without a mediator, (c)results for RF with a mediator (d) results for XGB with a mediator.}
  \label{ecdf_sample_size}
\end{figure}

\begin{figure}[H]
  \centering
  \includegraphics[width=0.88\textwidth]{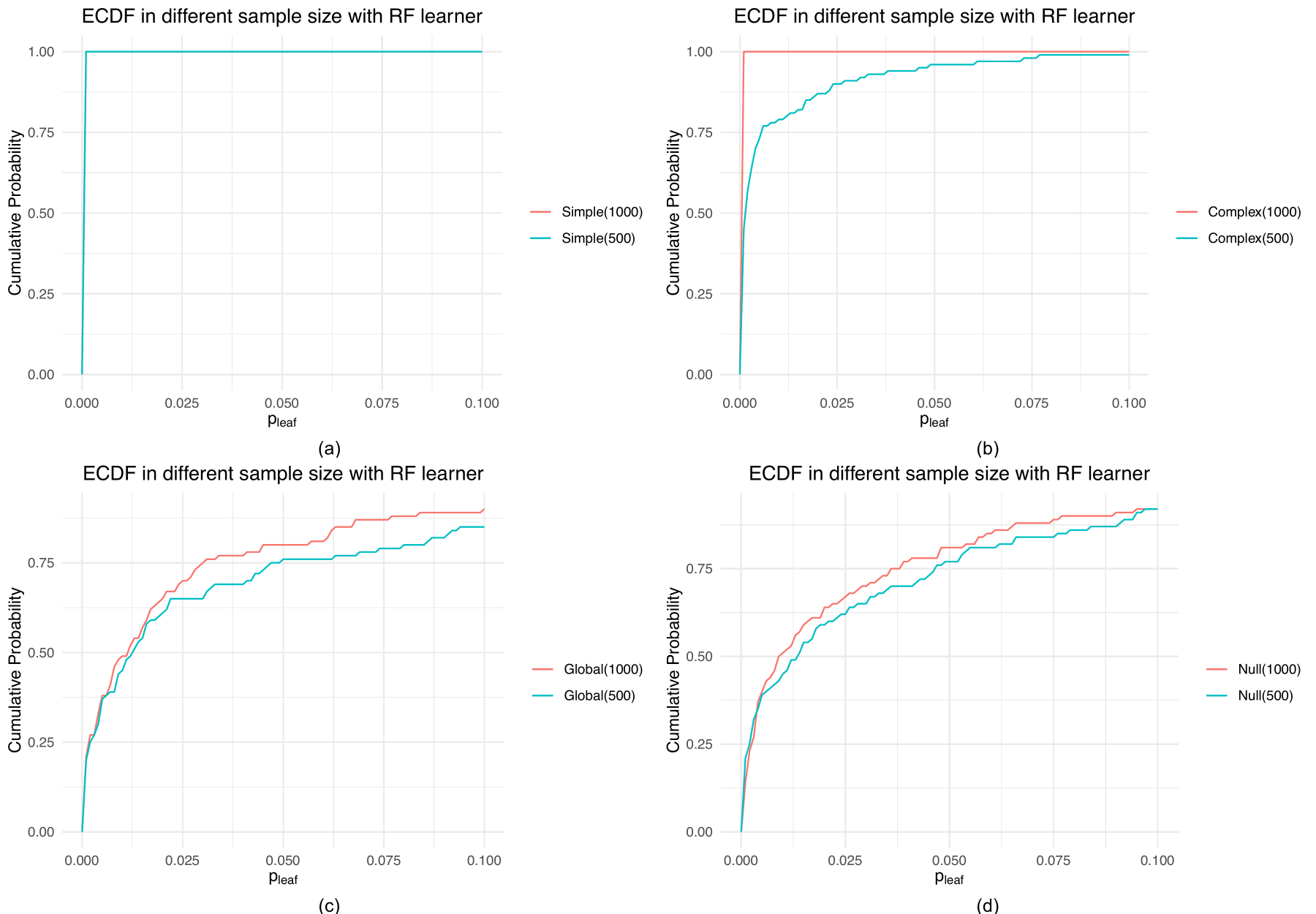}
  \caption{
  Comparison of \(p_{leaf}\) ECDFs under four scenarios without a mediator, using RF as base learner with sample sizes of \(1000\) and \(500\).
  (a)results for Simple scenario,
  (b) results for Complex scenario, (c)results for Global scenario (d) results for Null scenario.}
  \label{rf_41_p_no_mediator}
\end{figure}

\begin{figure}[H]
  \centering
  \includegraphics[width=0.88\textwidth]{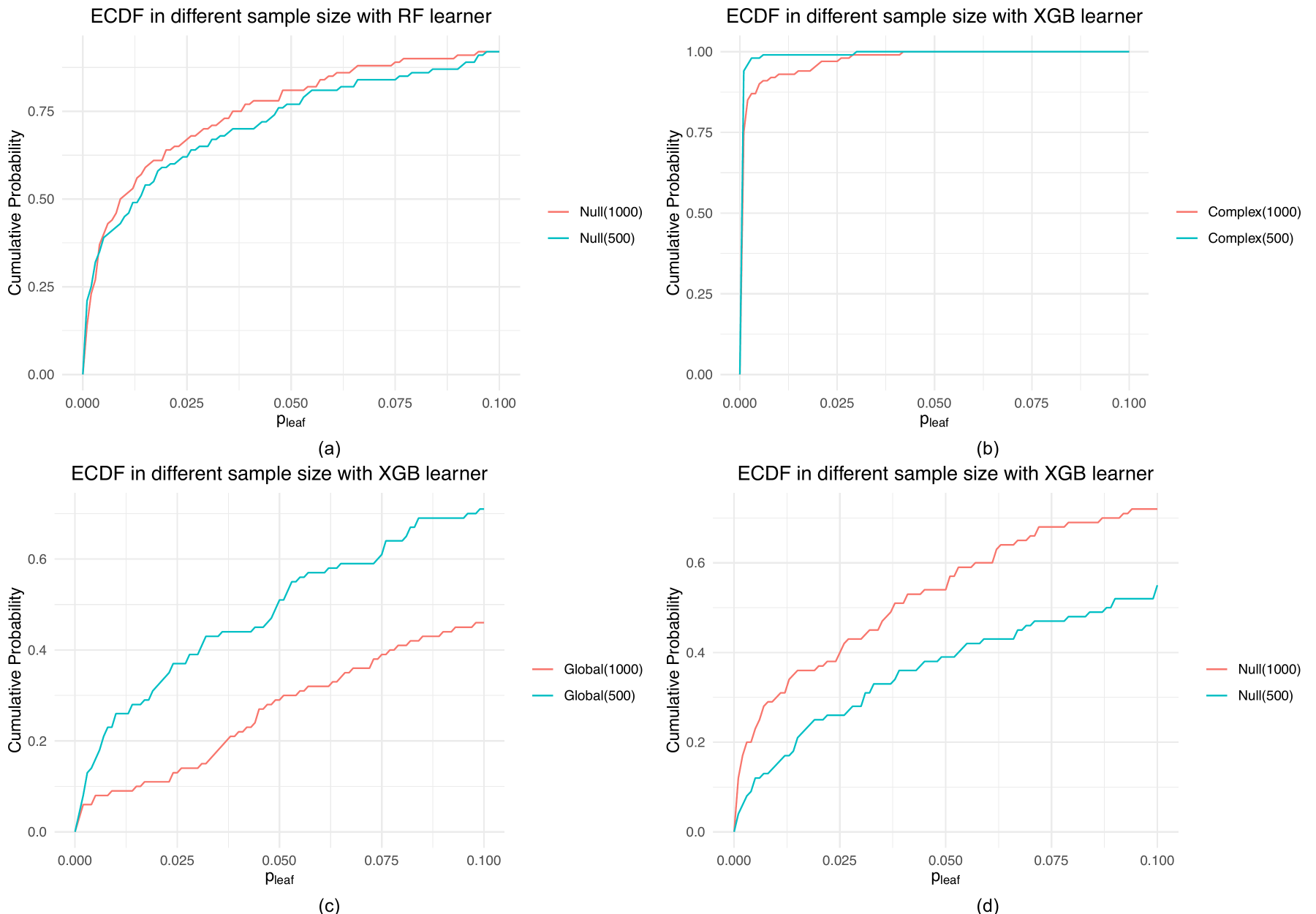}
  \caption{Comparison of \(p_{leaf}\) ECDFs under four scenarios without a mediator, using XGB as base learner with sample sizes of \(1000\) and \(500\).(a)results for Simple scenario,
  (b) results for Complex scenario, (c)results for Global scenario (d) results for Null scenario.}
  \label{xgb_41_p_no_mediator}
\end{figure}

\begin{figure}[H]
  \centering
  \includegraphics[width=1\textwidth]{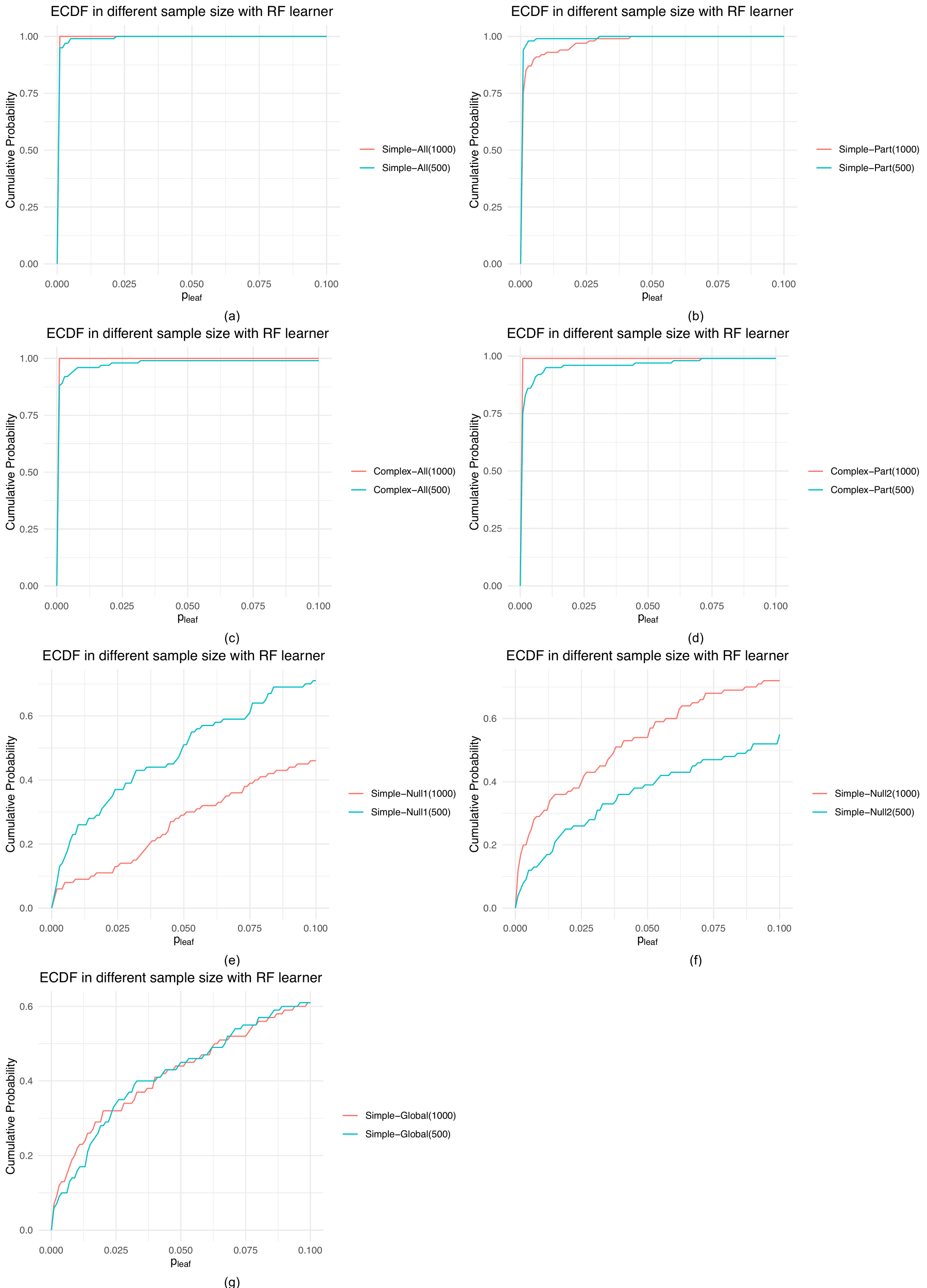}
  \caption{Comparison of \(p_{leaf}\) ECDFs under seven scenarios with a mediator, using RF as base learner with sample sizes of \(1000\) and \(500\).
  (a)results for Simple-All scenario,
  (b) results for Simple-Part scenario, (c)results for Complex-All scenario, (d) results for Complex-Part scenario,(e) results for Simple-Null1 scenario,(f) results for Simple-Null2 scenario.(g) results for Simple-Global scenario.}
  \label{rf_71_p_mediator}
\end{figure}

\begin{figure}[H]
  \centering
  \includegraphics[width=1\textwidth]{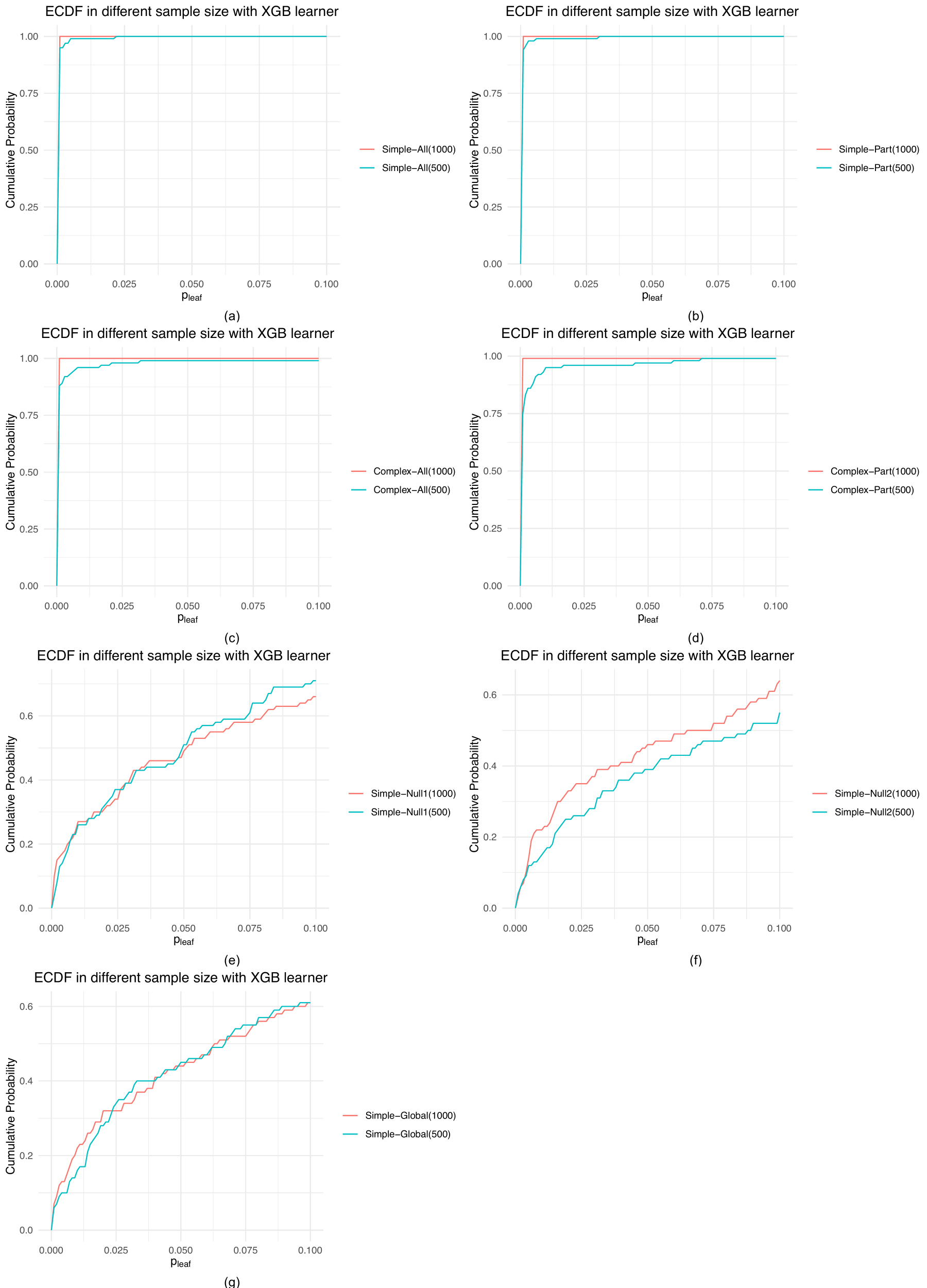}
  \caption{Comparison of \(p_{leaf}\) ECDFs under seven scenarios with a mediator, using XGB as base learner with sample sizes of \(1000\) and \(500\).
  (a)results for Simple-All scenario,
  (b) results for Simple-Part scenario, (c)results for Complex-All scenario, (d) results for Complex-Part scenario,(e) results for Simple-Null1 scenario,(f) results for Simple-Null2 scenario.(g) results for Simple-Global scenario.}
  \label{xgb_71_p_mediator}
\end{figure}

\subsection{K-Means}
\label{Kmeans_algorithm}
For the benchmark method K-means, clustering is performed directly on the data, with the number of clusters preset to range from \(2\) to \(5\). The \(p_{leaf}\) is then estimated using the same approach as in the M-learner framework, followed by a calibration procedure.

When model without mediators, the Appendix Table \ref{Hit_no_mediator_kmeans} and \ref{hit_kmeans_nomediator_dis} indicate the K-Means fails to effectively identify heterogeneous treatment subtypes and also struggles to distinguish non-heterogeneous scenarios.


When model with a mediator, 
the Appendix Table \ref{Hit_mediator_kmeans_dis}, \ref{kmeans_mediator_dis} and \ref{kmeans_mediator_dis_heter}
also indicate the K-Means fails to effectively identify effective subtypes and also struggles to distinguish non-heterogeneous scenarios. K-means completely fails to identify the Simple-Global scenario, which lacks treatment heterogeneity.

\begin{figure}[H]
  \centering
    \includegraphics[width=1\linewidth]{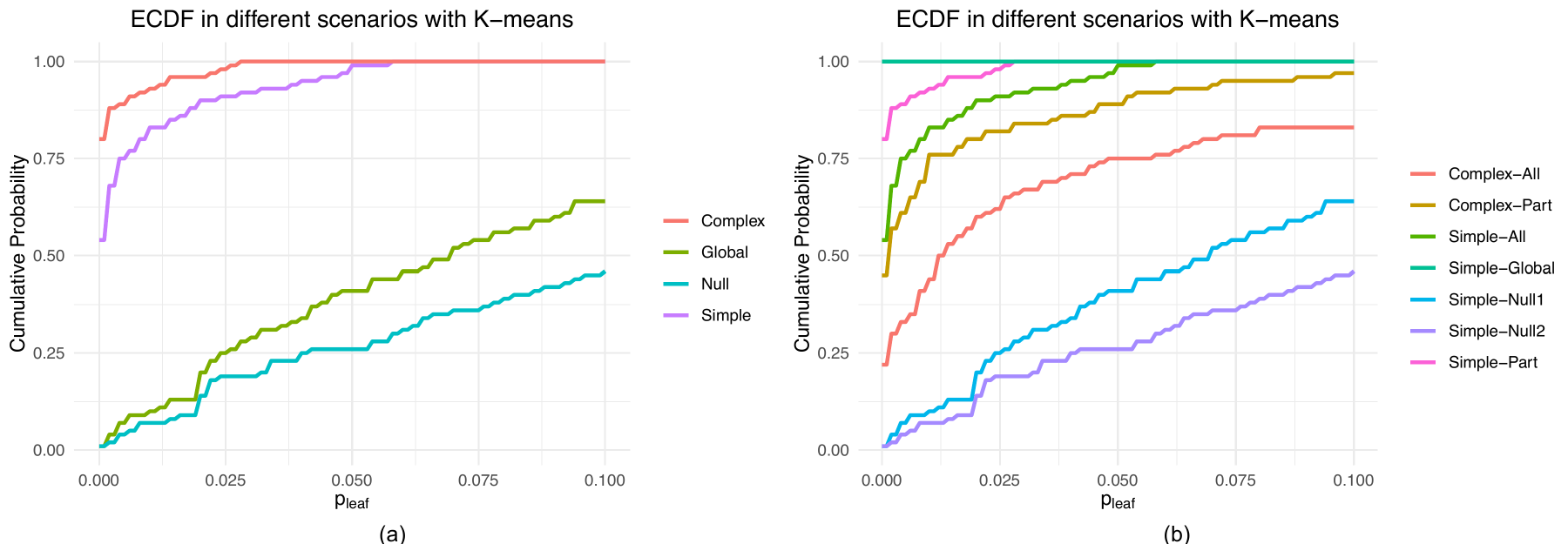}
  \caption{
  Empirical cumulative distribution functions (ECDF) of \(p_{leaf}\) under four different scenarios, using K-means. These results illustrate the sensitivity of each base learner to underlying treatment effect heterogeneity across varying levels of complexity.(a)results for K-means without a mediator, (b) results for K-means with a mediator.}
  \label{calibration_kmeans}
\end{figure}

The Appendix Figure \ref{calibration_kmeans} shows the ECDFs of K-means method.


\begin{table}[H]
\caption{The table summarizes the correct covariates in profiles in each scenario for K-means method in no mediators model. \(X_1/X_2\):Final profile contain \(X_1/X_2\), \(X_1\&X_2\): final profile contain both \(X_1\) and \(X_2\),\(X_1\) and \(X_2\) are the two variables associated with treatment effect heterogeneity. Each value denotes the count of occurrences across 100 simulations. }
    \centering
    \begin{tabular}{>{\centering\arraybackslash}p{2cm} 
                    >{\centering\arraybackslash}p{1cm}  
                    >{\centering\arraybackslash}p{1cm}
                    >{\centering\arraybackslash}p{1.5cm}
                    }
        \hline
                Covariates&\(X_1\)&\(X_2\)&\(X_1\&X_2\)\\
                \hline
                Simple&46&33&17\\
                Complex &22&10&4\\
                Global &65&48&24\\
                Null &3&2&1\\
                \hline
		\end{tabular}
        \label{Hit_no_mediator_kmeans}
\end{table}

\begin{table}[H]
    \centering
    \caption{ 
        The table summarizes the distribution of the number of variables included in the selected decision trees across 100 simulation runs. Each value represents the frequency with which a specific number of variables was selected over the course of the experiments. The performance of K-means was systematically evaluated under four distinct experimental scenarios. Selecting zero covariates in a given experimental run indicates that the decision tree’s classification was not supported by the calibration process and was therefore rejected.
        }
    \begin{tabular}{>{\centering\arraybackslash}p{3.0cm} 
                    >{\centering\arraybackslash}p{0.4cm}
                    >{\centering\arraybackslash}p{0.4cm}
                    >{\centering\arraybackslash}p{0.4cm}
                    >{\centering\arraybackslash}p{0.4cm}
                    >{\centering\arraybackslash}p{0.4cm}
                    >{\centering\arraybackslash}p{0.4cm}
                    }
        \hline
        Number of covariates&0&1&2&3&4&\(\geq\)5\\
        \hline
         Simple&33&6&18&19&15&9\\
         Complex&65&2&14&10&7&2\\
         Global&7&4&19&32&27&11\\
         Null&90&0&2&3&4&1\\
        \hline  
		\end{tabular}
        \label{hit_kmeans_nomediator_dis}
\end{table}

\begin{table}[H]
\caption{The table summarizes the correct covariates in profiles in each scenario for K-means method in the mediator model. \(X_1/X_2\):Final profile contain \(X_1/X_2\), \(X_1\&X_2\): final profile contain both \(X_1\) and \(X_2\),\(X_1\) and \(X_2\) are the two variables associated with treatment effect heterogeneity. Each value denotes the count of occurrences across 100 simulations. }
    \centering
    \begin{tabular}{>{\centering\arraybackslash}p{2.5cm} 
                    >{\centering\arraybackslash}p{1cm}  
                    >{\centering\arraybackslash}p{1cm}
                    >{\centering\arraybackslash}p{1.5cm}
                    }
        \hline
                Covariates&\(X_1\)&\(X_2\)&\(X_1\&X_2\)\\
                \hline
                Simple-All&53&36&17\\
                Simple-Part&57&38&18\\
                Complex-All &35&24&11\\
                Complex-Part &47&33&15\\
                Simple-Null1 &5&4&1\\
                Simple-Null2 &1&5&1\\
                Simple-Global &37&26&11\\
                \hline
		\end{tabular}
        \label{Hit_mediator_kmeans_dis}
\end{table}

\begin{table}[H]
    \centering
    \caption{ 
        The table summarizes the distribution of the number of variables included in the selected decision trees across 100 simulation runs. Each value represents the frequency with which a specific number of variables was selected over the course of the experiments. The performance of K-means was systematically evaluated under seven distinct experimental scenarios. Selecting zero covariates in a given experimental run indicates that the decision tree’s classification was not supported by the calibration process and was therefore rejected.
        }
    \begin{tabular}{>{\centering\arraybackslash}p{3.1cm} 
                    >{\centering\arraybackslash}p{0.4cm} 
                    >{\centering\arraybackslash}p{0.4cm}
                    >{\centering\arraybackslash}p{0.4cm}
                    >{\centering\arraybackslash}p{0.4cm}
                    >{\centering\arraybackslash}p{0.4cm}
                    >{\centering\arraybackslash}p{0.4cm}
                    }
        \hline
        Number of covariates&0&1&2&3&4&\(\geq\)5\\
        \hline
         Simple-All&12&6&23&25&22&12\\
         Simple-Part&4&8&25&29&22&12\\
         Complex-All&43&3&14&18&13&9\\
         Complex-Part&20&6&23&23&16&12\\
         Simple-Null1&87&0&3&3&5&2\\
         Simple-Null2&91&0&1&3&4&1\\
         Simple-Global&0&5&20&31&25&9\\
        \hline  
		\end{tabular}
        \label{kmeans_mediator_dis}
\end{table}

\begin{table}[H]
    \centering
    \caption{ 
    Mediation proportions and sample sizes within heterogeneous treatment effect regions across scenarios with K-means method.Standard deviations are shown in parentheses. The mediation proportion is calculated by R package "mediation". "For each set of results, the subtype exhibiting the lowest p-value for the mediation proportion is designated as the region of interest. Within this region, we estimate the mediation proportion, total treatment effect, and indirect treatment effect.
        }
    \begin{tabular}{>{\centering\arraybackslash}p{2.0cm} 
                    >{\centering\arraybackslash}p{1.5cm}
                    >{\centering\arraybackslash}p{1.5cm}
                    >{\centering\arraybackslash}p{1.5cm}
                    >{\centering\arraybackslash}p{1.5cm}
                    }
        \hline
        &\multicolumn{2}{c}{True}&\multicolumn{2}{c}{K-means}\\
        \hline
        Scenario&N&Med Prop&N&Med Prop\\
        \hline
        Simple-All&251(13)&1.37(0.09)&264(121)&1.21(0.11)\\
        Simple-Part&251(13)&0.76(0.03)&278(127)&0.80(0.05)\\
        Complex-All&251(13)&1.54(0.30)&265(118)&1.30(0.23)\\
        Complex-Part&251(13)&0.85(0.08)&284(122)&1.03(0.14)\\
        \hline
		\end{tabular}
        \label{kmeans_mediator_dis_heter}
\end{table}

\subsection{Sensitivity analysis}
\label{sensitivity}

\subsubsection{Noise}
To evaluate the robustness of the proposed method under varying levels of noise, we introduced additive Gaussian noise with different variances. The noise levels were categorized as follows: 
\begin{itemize}
    \item low noise (original setting): \(N(0,0.01)\);
    \item moderate noise: \(N(0,0.1)\);
    \item high noise: \(N(0,1)\)
.   
\end{itemize}
All other experimental settings were kept identical to Simple-All and Complex-All in  \ref{setting_with}. For both the moderate and high noise settings, calibration is performed using the Simple-Null2 distribution obtained under the low noise scenario (To simulate real-world data characteristics, we perform calibration using data with a low level of noise).
What's more, we want to evaluate the error of the selected heterogeneous region.
Here, we define the threshold error, for each run, we extracted the decision threshold defining the heterogeneous region and computed the average of the squared threshold values. If a variable was not selected or the threshold exceeded 1, we assigned a value of 1.

\begin{figure}[H]
  \centering
    \includegraphics[width=0.9\linewidth]{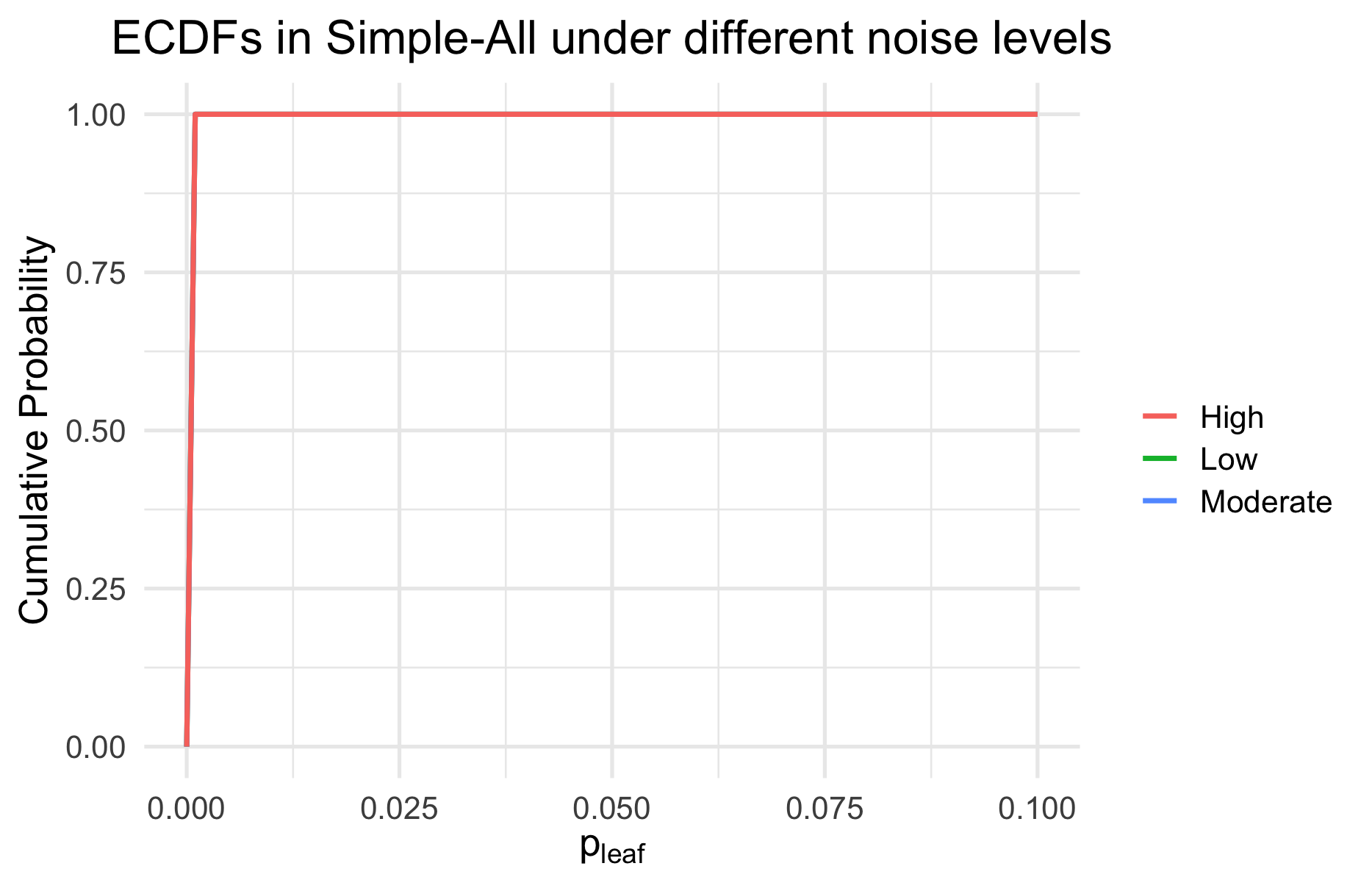}
  \caption{
  Empirical cumulative distribution functions (ECDF) of \(p_{leaf}\) under different noises levels in Simple-All scenario. These results illustrate the sensitivity of each base learner to underlying treatment effect heterogeneity across varying levels of complexity.}
  \label{calibration_noise_sc_all}
\end{figure}

From Table \ref{Hit_mediator_noise_dis_simple_all}, \ref{noise_sc1_mediator_dis},\ref{boundary_noise_simple} and Figure \ref{calibration_noise_sc_all},  suggest that noise level does have some effect on the model's performance in the Simple-All scenario. In particular, performance tends to degrade slightly under higher levels of noise.

\begin{figure}[H]
  \centering
    \includegraphics[width=0.9\linewidth]{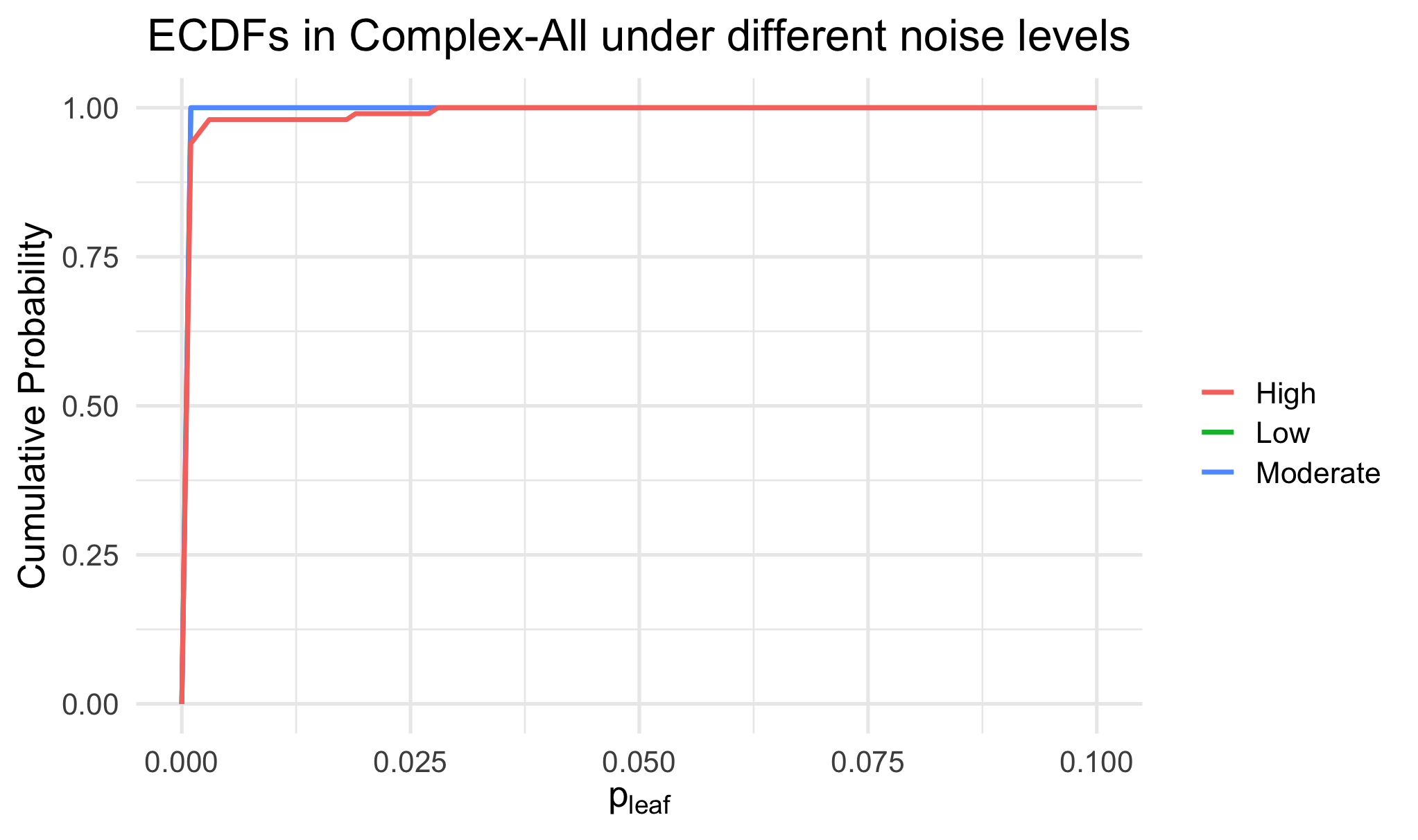}
  \caption{
  Empirical cumulative distribution functions (ECDF) of \(p_{leaf}\) under different noises levels in Complex-All scenario. These results illustrate the sensitivity of each base learner to underlying treatment effect heterogeneity across varying levels of complexity.}
  \label{calibration_noise_complex_all}
\end{figure}

From Table \ref{Hit_mediator_noise_dis_complex_all}, \ref{noise_sc6_mediator_dis_complex_all},\ref{boundary_noise_complex} and Figure \ref{calibration_noise_complex_all}, suggest that noise level does have some effect on the model's performance in the Complex-All scenario. Similar with Simple-All scenario, performance tends to degrade slightly under higher levels of noise.

These experimental findings provide insights into the effect of noise on model performance. While noise introduces some variability, its overall impact is limited. Notably, under high-noise conditions, the learner based on XGBoost even outperforms the Random Forest learner under low-noise settings.

\begin{table}[H]
\caption{The table summarizes the correct covariates in profiles in  scenario Simple-All for different noise levels with XGB learner. \(X_1/X_2\):Final profile contain \(X_1/X_2\), \(X_1\&X_2\): final profile contain both \(X_1\) and \(X_2\),\(X_1\) and \(X_2\) are the two variables associated with treatment effect heterogeneity. Each value denotes the count of occurrences across 100 simulations. }
    \centering
    \begin{tabular}{>{\centering\arraybackslash}p{2.5cm} 
                    >{\centering\arraybackslash}p{1cm}  
                    >{\centering\arraybackslash}p{1cm}
                    >{\centering\arraybackslash}p{1.5cm}
                    }
        \hline
                Covariates&\(X_1\)&\(X_2\)&\(X_1\&X_2\)\\
                \hline
                Low&100&100&100\\
                Moderate&96&98&95\\
                High &91&94&85\\
                \hline
		\end{tabular}
        \label{Hit_mediator_noise_dis_simple_all}
\end{table}

\begin{table}[H]
\caption{The table summarizes the correct covariates in profiles in  scenario Complex-All for different noise levels with XGB learner. \(X_1/X_2\):Final profile contain \(X_1/X_2\), \(X_1\&X_2\): final profile contain both \(X_1\) and \(X_2\),\(X_1\) and \(X_2\) are the two variables associated with treatment effect heterogeneity. Each value denotes the count of occurrences across 100 simulations. }
    \centering
    \begin{tabular}{>{\centering\arraybackslash}p{2.5cm} 
                    >{\centering\arraybackslash}p{1cm}  
                    >{\centering\arraybackslash}p{1cm}
                    >{\centering\arraybackslash}p{1.5cm}
                    }
        \hline
                Covariates&\(X_1\)&\(X_2\)&\(X_1\&X_2\)\\
                \hline
                Low&98&99&97\\
                Moderate&100&100&100\\
                High &83&87&72\\
                \hline
		\end{tabular}
        \label{Hit_mediator_noise_dis_complex_all}
\end{table}

\begin{table}[H]
\caption{The table summarizes the boundary osf selected heterogeneous regions in profiles in  scenario Simple-All for different noise levels with XGB learner. Values are the mean threshold errors over 100 replications, with standard deviations shown in parentheses. }
    \centering
    \begin{tabular}{>{\centering\arraybackslash}p{2.5cm} 
                    >{\centering\arraybackslash}p{2cm}  
                    >{\centering\arraybackslash}p{2cm}
                    }
        \hline
                Covariates&\(X_1\)&\(X_2\)\\
                \hline
                Low&0.056(0.194)&0.059(0.162)\\
                Moderate&0.043(0.173)&0.047(0.153)\\
                High &0.111(0.285)&0.106(0.289)\\
                \hline
		\end{tabular}
        \label{boundary_noise_simple}
\end{table}

\begin{table}[H]
\caption{The table summarizes the boundary osf selected heterogeneous regions in profiles in  scenario Complex-All for different noise levels with XGB learner. Values are the mean threshold errors over 100 replications, with standard deviations shown in parentheses. }
    \centering
    \begin{tabular}{>{\centering\arraybackslash}p{2.5cm} 
                    >{\centering\arraybackslash}p{2cm}  
                    >{\centering\arraybackslash}p{2cm}
                    }
        \hline
                Covariates&\(X_1\)&\(X_2\)\\
                \hline
                Low&0.425(0.489)&0.475(0.481)\\
                Moderate&0.372(0.483)&0.531(0.529)\\
                High &0.419(0.478)&0.561(0.475)\\
                \hline
		\end{tabular}
        \label{boundary_noise_complex}
\end{table}

\begin{table}[H]
    \centering
    \caption{ 
        The table summarizes the distribution of the number of variables included in the selected decision trees across 100 simulation runs with different noises levels in Simple-All scenario. Each value represents the frequency with which a specific number of variables was selected over the course of the experiments.  Selecting zero covariates in a given experimental run indicates that the decision tree’s classification was not supported by the calibration process and was therefore rejected.
        }
    \begin{tabular}{>{\centering\arraybackslash}p{3.1cm} 
                    >{\centering\arraybackslash}p{0.4cm} 
                    >{\centering\arraybackslash}p{0.4cm}
                    >{\centering\arraybackslash}p{0.4cm}
                    >{\centering\arraybackslash}p{0.4cm}
                    >{\centering\arraybackslash}p{0.4cm}
                    >{\centering\arraybackslash}p{0.4cm}
                    }
        \hline
        Number of covariates&0&1&2&3&4&\(\geq\)5\\
        \hline
         Low&0&0&58&33&9&0\\
         Moderate&1&2&79&15&3&0\\
         High&0&12&63&18&7&0\\
         
        \hline  
		\end{tabular}
        \label{noise_sc1_mediator_dis}
\end{table}

\begin{table}[H]
    \centering
    \caption{ 
        The table summarizes the distribution of the number of variables included in the selected decision trees across 100 simulation runs with different noises levels in Complex-All scenario. Each value represents the frequency with which a specific number of variables was selected over the course of the experiments.  Selecting zero covariates in a given experimental run indicates that the decision tree’s classification was not supported by the calibration process and was therefore rejected.
        }
    \begin{tabular}{>{\centering\arraybackslash}p{3.1cm} 
                    >{\centering\arraybackslash}p{0.4cm} 
                    >{\centering\arraybackslash}p{0.4cm}
                    >{\centering\arraybackslash}p{0.4cm}
                    >{\centering\arraybackslash}p{0.4cm}
                    >{\centering\arraybackslash}p{0.4cm}
                    >{\centering\arraybackslash}p{0.4cm}
                    }
        \hline
        Number of covariates&0&1&2&3&4&\(\geq\)5\\
        \hline
         Low&0&0&58&33&9&0\\
         Moderate&0&0&19&47&34&0\\
         High&2&6&33&40&16&3\\
         
        \hline  
		\end{tabular}
        \label{noise_sc6_mediator_dis_complex_all}
\end{table}

\begin{figure}[H]
  \centering
    \includegraphics[width=0.5\linewidth]{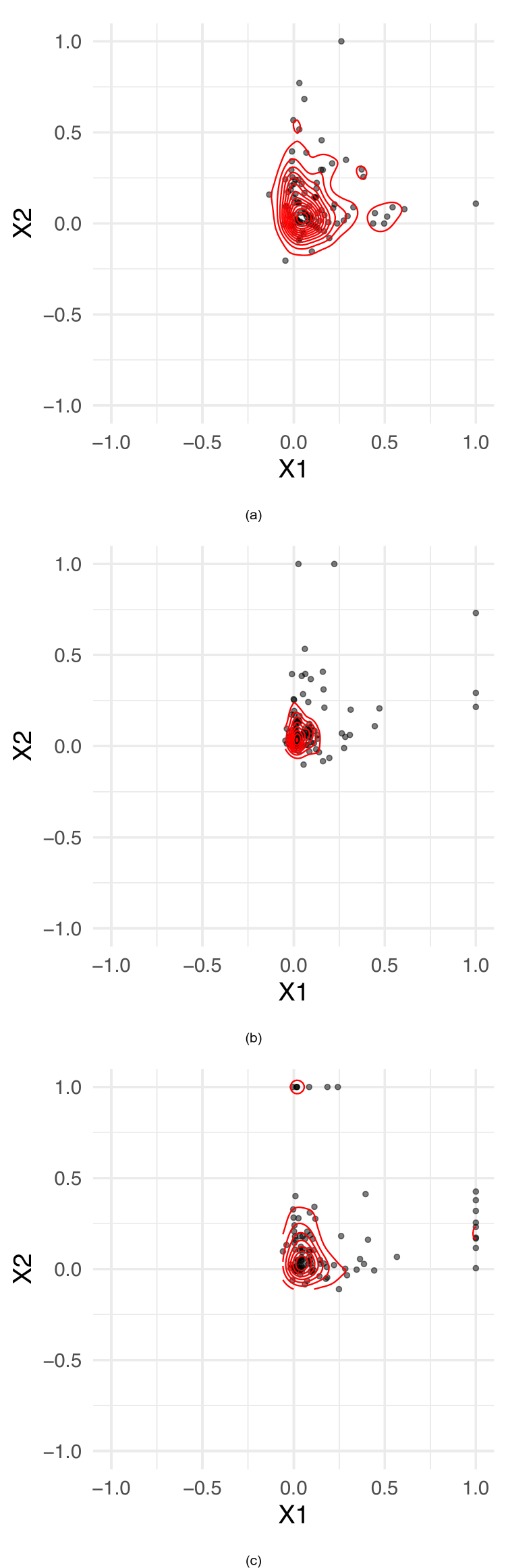}
  \caption{
  Threshold location distribution plot for Simple-All scenario. dot in the figure represents the threshold of heterogeneous region for 100 replications, red line represents the density line. If a variable was not selected or the threshold exceeded 1, we assigned a value of 1. (a) Low noise, (b) Moderate noise, (c) High noise.}
  \label{noise_expe1}
\end{figure}

\begin{figure}[H]
  \centering
    \includegraphics[width=0.5\linewidth]{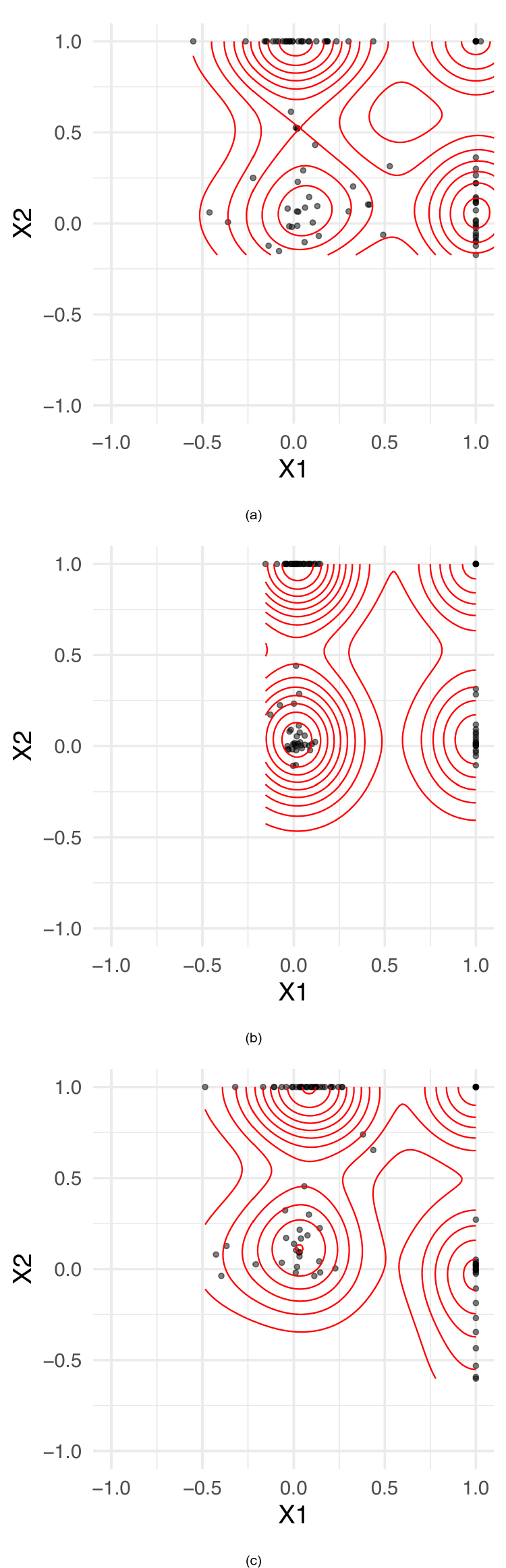}
  \caption{
  Threshold location distribution plot for Complex-All scenario. dot in the figure represents the threshold of heterogeneous region for 100 replications, red line represents the density line. If a variable was not selected or the threshold exceeded 1, we assigned a value of 1. (a) Low noise, (b) Moderate noise, (c) High noise.}
  \label{noise_expe6}
\end{figure}


\subsubsection{dimension of projection}

Here, we evaluate the robustness of the dimension of projection. 
t-SNE only support reduce dimension to \(2\) or \(3\). Here we compare the results in Simple-All and Complex-All scenario with different dimension of projections. For UMAP, it support reduce to higher dimension, we compare the results with \(2, 5\) and \(10\) in Simple-All and Complex-All scenario.

As shown in the Table \ref{hit_projection_dimension}, the performance of t-SNE appears insensitive to the choice of projection dimension, while UMAP exhibits a clear dependence on it. In particular, UMAP achieves optimal performance when the projection dimension is set to 5, outperforming both the 2- and 10-dimensional settings. Furthermore, even the best t-SNE result is consistently outperformed by UMAP under its optimal configuration.

\begin{table}[h]
\caption{The table summarizes the correct covariates in profiles in each scenario. \(X_1/X_2\):Final profile contain \(X_1/X_2\), \(X_1\&X_2\): final profile contain both \(X_1\) and \(X_2\), \(X_1\) and \(X_2\) are the two variables associated with treatment effect heterogeneity. Each value denotes the count of occurrences across 100 simulations. }
    \centering
    \begin{tabular}{>{\centering\arraybackslash}p{2.3cm} 
                    >{\centering\arraybackslash}p{1cm}  
                    >{\centering\arraybackslash}p{1cm}
                    >{\centering\arraybackslash}p{1.5cm}
                    >{\centering\arraybackslash}p{0.01cm}
                    >{\centering\arraybackslash}p{1cm}  
                    >{\centering\arraybackslash}p{1cm}
                    >{\centering\arraybackslash}p{1.5cm}
                    }
        \hline
        &\multicolumn{3}{c}{Simple-All}&&\multicolumn{3}{c}{Complex-All}\\
        \hline
                Covariates&\(X_1\)&\(X_2\)&\(X_1\&X_2\)&&\(X_1\)&\(X_2\)&\(X_1\&X_2\)\\
                \hline
                t-SNE(2)&100&100&100&&100&100&100\\
                t-SNE(3)&100&100&100&&100&100&100\\
                UMAP(2)&97&98&96&&91&86&82\\
                UMAP(5)&99&100&99&&96&92&91\\
                UMAP(10)&94&94&89&&89&84&79\\
                \hline
		\end{tabular}
        \label{hit_projection_dimension}
\end{table}

\subsubsection{Number of maximum clusters}
To investigate the impact of the maximum number of clusters on the results, we evaluate the effect of setting the maximum cluster number to 2, 5, and 10 in both the Simple-All and Complex-All scenarios. From Table \ref{hit_clusters} and \ref{mediator_dis_clusters}, we observe that the maximum number of clusters 
\(K\) influences the results to some extent, but this effect becomes negligible once 
\(K\) exceeds a certain threshold. Therefore, in practice, we recommend choosing an appropriate value of 
\(K\) based on computational efficiency. To ensure optimal performance, we suggest selecting the largest feasible 
\(K\) within resource constraints.

\begin{table}[h]
\caption{The table summarizes the correct covariates in profiles in Simple-All and Complex-All scenario for different maximum clusters \(K\). \(X_1/X_2\):Final profile contain \(X_1/X_2\), \(X_1\&X_2\): final profile contain both \(X_1\) and \(X_2\), \(X_1\) and \(X_2\) are the two variables associated with treatment effect heterogeneity. Each value denotes the count of occurrences across 100 simulations. }
    \centering
    \begin{tabular}{>{\centering\arraybackslash}p{2.3cm} 
                    >{\centering\arraybackslash}p{1cm}  
                    >{\centering\arraybackslash}p{1cm}
                    >{\centering\arraybackslash}p{1.5cm}
                    >{\centering\arraybackslash}p{0.01cm}
                    >{\centering\arraybackslash}p{1cm}  
                    >{\centering\arraybackslash}p{1cm}
                    >{\centering\arraybackslash}p{1.5cm}
                    }
        \hline
        &\multicolumn{3}{c}{Simple-All}&&\multicolumn{3}{c}{Complex-All}\\
        \hline
                Covariates&\(X_1\)&\(X_2\)&\(X_1\&X_2\)&&\(X_1\)&\(X_2\)&\(X_1\&X_2\)\\
                \hline
                \(K=2\)&87&83&75&&81&87&72\\
                \(K=5\)&100&100&100&&98&99&97\\
                \(K=10\)&100&100&100&&100&100&100\\
                \hline
		\end{tabular}
        \label{hit_clusters}
\end{table}

\begin{table}[H]
    \centering
    \caption{ 
        The table summarizes the distribution of the number of variables included in the selected decision trees across 100 simulation runs when the sample size is \(500\). Each value represents the frequency with which a specific number of variables was selected over the course of the experiments. The performance of Random Forest and XGBoost as base learners was systematically evaluated under seven distinct experimental scenarios. Selecting zero covariates in a given experimental run indicates that the decision tree’s classification was not supported by the calibration process and was therefore rejected.
        }
    \begin{tabular}{>{\centering\arraybackslash}p{3.3cm} 
                    >{\centering\arraybackslash}p{0.4cm}  
                    >{\centering\arraybackslash}p{0.4cm}
                    >{\centering\arraybackslash}p{0.4cm}
                    >{\centering\arraybackslash}p{0.4cm}
                    >{\centering\arraybackslash}p{0.4cm}
                    >{\centering\arraybackslash}p{0.4cm}
                    >{\centering\arraybackslash}p{0.01cm}
                    >{\centering\arraybackslash}p{0.4cm}
                    >{\centering\arraybackslash}p{0.4cm}
                    >{\centering\arraybackslash}p{0.4cm}
                    >{\centering\arraybackslash}p{0.4cm}
                    >{\centering\arraybackslash}p{0.4cm}
                    >{\centering\arraybackslash}p{0.4cm}
                    }
        \hline
        Base Learner&\multicolumn{6}{c}{Simple-All}&&\multicolumn{6}{c}{Complex-All}\\
        \hline
        Number of covariates&0&1&2&3&4&\(\geq\)5&&0&1&2&3&4&\(\geq\)5\\
        \hline
         \(K=2\)&5&12&66&14&3&0&&4&12&54&21&8&1\\
         \(K=5\)&3&1&46&39&11&0&&0&1&23&43&33&0\\
         \(K=10\)&37&6&16&23&13&0&&0&0&9&38&53&0\\
         
        \hline  
		\end{tabular}
        \label{mediator_dis_clusters}
\end{table}

\subsection{Comparison of projection methods}
\label{projection}
The core idea of our M-learner algorithm is that individuals with similar treatment effects should be considered more similar. Based on this intuition, we construct a pairwise treatment effect distance matrix that reflects heterogeneity between any two individuals. We then use t-SNE to project this matrix into a Euclidean space for clustering.

t-SNE (similarly UMAP) is particularly suitable here because it is designed to preserve local pairwise similarities during projection, which aligns with our goal of preserving treatment effect similarity. Unlike PCA, which performs linear projections and assumes Euclidean structure in the original space, our distance matrix is derived from treatment effect heterogeneity—not raw features—making non-linear methods like t-SNE or UMAP more appropriate. Notably, in our study, t-SNE directly operates on pairwise distances, which further justifies its application. In our study, t-SNE is employed to project the treatment effect distance matrix onto a low-dimensional Euclidean space while preserving local neighborhood structures. This approach facilitates the effective identification of potential subgroups, which are subsequently extracted through clustering.

Moreover, we also conducted sensitivity analyses using alternative projection methods such as UMAP(the goal in this step is not to reduce dimensionality per se, but rather to project the treatment distance matrix into a space that preserves their relative distances for downstream clustering. PCA focuses on preserving global variance rather than local or relational structure, it is not suitable for this purpose), and the results remained qualitatively consistent. 

In this section, we compare the results for two methods. In this experiment, we adopted XGBoost as the sole base learner. The configurations for both the without mediator and with mediator scenarios followed those of Experiment \ref{setting_no} and \ref{setting_with}, with the only difference being the choice of projection method.

\begin{figure}[H]
  \centering
    \includegraphics[width=1\linewidth]{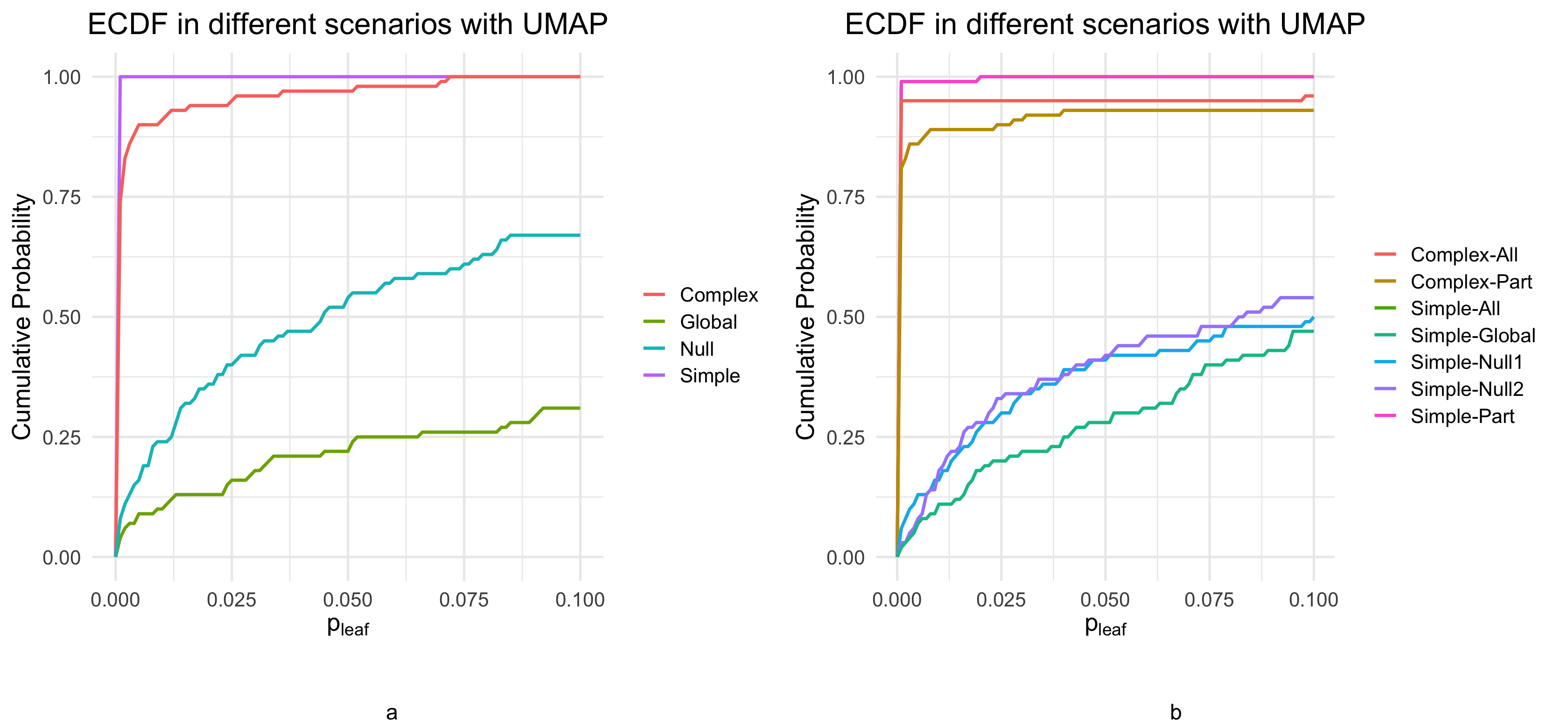}
  \caption{
  Empirical cumulative distribution functions (ECDF) of \(p_{leaf}\) for projection comparison.(a)results for UMAP without a mediator, (b) results for UMAP with a mediator.}
  \label{calibration_umap}
\end{figure}

From Table \ref{Hit_no_mediator_umap}, \ref{no_mediator_dis_umap} and Figure \ref{calibration_umap} results, we can think there is little difference between two projection methods when there is no mediators.
However, from the results from Table \ref{Hit_mediator_umap}, \ref{mediator_dis_umap} and Figure \ref{calibration_umap} when there exists a mediator, UMAP perform worse than t-SNE in complex scenarios. So we choose t-SNE as the projection method.

\begin{table}[h]
\caption{The table summarizes the correct covariates in profiles in each scenario. \(X_1/X_2\):Final profile contain \(X_1/X_2\), \(X_1\&X_2\): final profile contain both \(X_1\) and \(X_2\),\(X_1\) and \(X_2\) are the two variables associated with treatment effect heterogeneity. Each value denotes the count of occurrences across 100 simulations. }
    \centering
    \begin{tabular}{>{\centering\arraybackslash}p{3cm} 
                    >{\centering\arraybackslash}p{1cm}  
                    >{\centering\arraybackslash}p{1cm}
                    >{\centering\arraybackslash}p{1.5cm}
                    >{\centering\arraybackslash}p{0.01cm}
                    >{\centering\arraybackslash}p{1cm}  
                    >{\centering\arraybackslash}p{1cm}
                    >{\centering\arraybackslash}p{1.5cm}
                    }
        \hline
        Projection method&\multicolumn{3}{c}{UMAP}&&\multicolumn{3}{c}{t-SNE}\\
        \hline
                Covariates&\(X_1\)&\(X_2\)&\(X_1\&X_2\)&&\(X_1\)&\(X_2\)&\(X_1\&X_2\)\\
                \hline
                Simple&100&100&100&&100&100&100\\
                Complex &72&74&68&&70&70&68\\
                Global &0&1&0&&0&2&0\\
                Null &2&2&1&&1&4&1\\
                \hline
		\end{tabular}
        \label{Hit_no_mediator_umap}
        
\end{table}

\begin{table}[H]
    \centering
    \caption{ 
        The table summarizes the distribution of the number of variables included in the selected decision trees across 100 simulation runs when there is no mediator and sample size is \(1000\). Each value represents the frequency with which a specific number of variables was selected over the course of the experiments. The base learner is XGBoost.
This table compares UMAP and t-SNE as two distinct projection methods under four distinct
experimental scenarios Selecting zero covariates in a given experimental run indicates that the decision tree’s classification was not supported by the calibration process and was therefore rejected.
        }
    \begin{tabular}{>{\centering\arraybackslash}p{3.0cm} 
                    >{\centering\arraybackslash}p{0.4cm}  
                    >{\centering\arraybackslash}p{0.4cm}
                    >{\centering\arraybackslash}p{0.4cm}
                    >{\centering\arraybackslash}p{0.4cm}
                    >{\centering\arraybackslash}p{0.6cm}
                    >{\centering\arraybackslash}p{0.001cm}
                    >{\centering\arraybackslash}p{0.4cm}
                    >{\centering\arraybackslash}p{0.4cm}
                    >{\centering\arraybackslash}p{0.4cm}
                    >{\centering\arraybackslash}p{0.4cm}
                    >{\centering\arraybackslash}p{0.6cm}
                    }
        \hline
        Projection method&\multicolumn{5}{c}{UMAP}&&\multicolumn{5}{c}{t-SNE}\\
        \hline
        Number of covariates&0&1&2&3&\(\geq\)4&&0&1&2&3&\(\geq\)4\\
        \hline
        Simple&0&0&62&28&10&&0&0&57&28&15\\
        Complex&21&1&15&29&14&&28&1&7&30&34\\
        Global&95&1&3&1&0&&98&0&2&0&0\\
        Null&90&2&4&4&0&&90&3&3&3&1\\
        \hline  
		\end{tabular}
        \label{no_mediator_dis_umap}
\end{table}

\begin{table}[h]
\caption{The table summarizes the correct covariates in profiles in each scenario. \(X_1/X_2\):Final profile contain \(X_1/X_2\), \(X_1\&X_2\): final profile contain both \(X_1\) and \(X_2\), \(X_1\) and \(X_2\) are the two variables associated with treatment effect heterogeneity. Each value denotes the count of occurrences across 100 simulations. }
    \centering
    \begin{tabular}{>{\centering\arraybackslash}p{2.3cm} 
                    >{\centering\arraybackslash}p{1cm}  
                    >{\centering\arraybackslash}p{1cm}
                    >{\centering\arraybackslash}p{1.5cm}
                    >{\centering\arraybackslash}p{0.01cm}
                    >{\centering\arraybackslash}p{1cm}  
                    >{\centering\arraybackslash}p{1cm}
                    >{\centering\arraybackslash}p{1.5cm}
                    }
        \hline
        &\multicolumn{3}{c}{UMAP}&&\multicolumn{3}{c}{t-SNE}\\
        \hline
                Covariates&\(X_1\)&\(X_2\)&\(X_1\&X_2\)&&\(X_1\)&\(X_2\)&\(X_1\&X_2\)\\
                \hline
                Simple-All&97&98&96&&100&100&100\\
                Simple-Part&100&99&99&&100&100&100\\
                Complex-All&91&86&82&&98&99&97\\
                Complex-Part&79&69&61&&95&97&93\\
                Simple-Null1&6&2&2&&8&4&2\\
                Simple-Null2&2&3&0&&2&1&0\\
                Simple-Global&1&3&1&&3&5&1\\
                \hline
		\end{tabular}
        \label{Hit_mediator_umap}
\end{table}

\begin{table}[H]
    \centering
    \caption{ 
        The table summarizes the distribution of the number of variables included in the selected decision trees across 100 simulation runs. Each value represents the frequency with which a specific number of variables was selected over the course of the experiments. The base learner is XGBoost. This table compares UMAP and t-SNE as two distinct projection methods under seven distinct experimental scenarios. Selecting zero covariates in a given experimental run indicates that the decision tree’s classification was not supported by the calibration process and was therefore rejected.
        }
    \begin{tabular}{>{\centering\arraybackslash}p{3.0cm} 
                    >{\centering\arraybackslash}p{0.4cm}  
                    >{\centering\arraybackslash}p{0.4cm}
                    >{\centering\arraybackslash}p{0.4cm}
                    >{\centering\arraybackslash}p{0.4cm}
                    >{\centering\arraybackslash}p{0.4cm}
                    >{\centering\arraybackslash}p{0.4cm}
                    >{\centering\arraybackslash}p{0.01cm}
                    >{\centering\arraybackslash}p{0.4cm}
                    >{\centering\arraybackslash}p{0.4cm}
                    >{\centering\arraybackslash}p{0.4cm}
                    >{\centering\arraybackslash}p{0.4cm}
                    >{\centering\arraybackslash}p{0.4cm}
                    >{\centering\arraybackslash}p{0.4cm}
                    }
        \hline
        Projection method&\multicolumn{6}{c}{UMAP}&&\multicolumn{6}{c}{t-SNE}\\
        \hline
        Number of covariates&0&1&2&3&4&\(\geq\)5&&0&1&2&3&4&\(\geq\)5\\
        \hline
         Simple-All&1&3&61&17&15&3&&0&0&58&33&9&0\\
         Simple-Part&0&1&68&23&8&0&&0&0&54&40&6&0\\
         Complex-All&5&6&25&42&22&0&&0&1&23&43&33&0\\
         Complex-Part&13&10&25&34&18&0&&1&1&14&49&35&0\\
         Simple-Null1&86&6&5&1&1&1&&83&1&8&6&1&1\\
         Simple-Null2&91&2&4&2&1&0&&90&5&3&2&0&0\\
         Simple-Global&90&2&7&1&0&0&&87&1&6&4&2&0\\
        \hline  
		\end{tabular}
        \label{mediator_dis_umap}
\end{table}

\subsection{Different clustering methods}

\begin{figure}[H]
  \centering
    \includegraphics[width=0.7\linewidth]{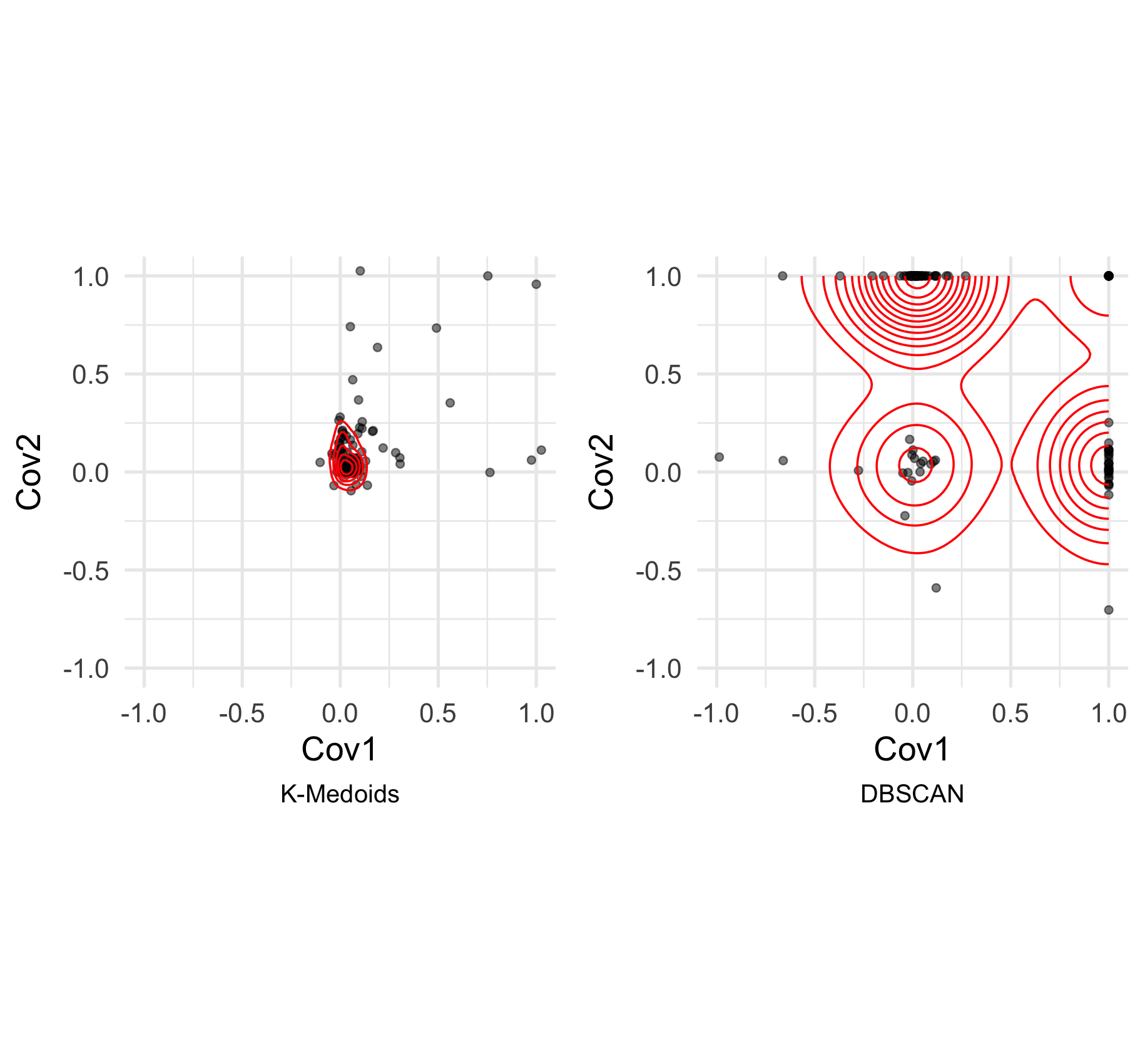}
  \caption{
  Empirical cumulative distribution functions (ECDF) of \(p_{leaf}\) for clustering comparison.(a)results for K-Medoids, (b) results for DBSCAN. Cov1 represnets "X1", Cov2 represnets "X2".}
  \label{border_clustering}
\end{figure}

We compare different clustering methods, such as K-Medoids and DBSCAN in Simple scenario (no mediator setting). DBSCAN does not depend on predefined numbers of clusters, we set eps is 1, minPts is 20 with R dbscan function. For K-Medoids, similar to K-Means, we preset the number of clusters to range from \(2\) to \(5\). Here, we compare the performance in simple scenario in non-mediated setting. from \ref{border_clustering}, K-Medoids method close to K-Means, and DBSCAN is not acceptable.

\subsection{Comparison in non-mediators setting}
\label{meta_learner}

When no mediators are present, our method is equivalent to the T-learner in the estimation of the CATTE. In this subsection, we compare the performance of several popular methods for subgroup identification (using the same clustering and balancing approach described later in this paper), including X-learner, R-learner, and a neural network-based method, TARNet. 
To ensure a fair comparison, all learner-based methods were implemented using XGBoost and estimated via the econml Python library. The TARNet model was implemented using the DragonNet architecture from the causalml library, with the propensity score weighting component removed.

\begin{figure}[H]
  \centering
    \includegraphics[width=0.7\linewidth]{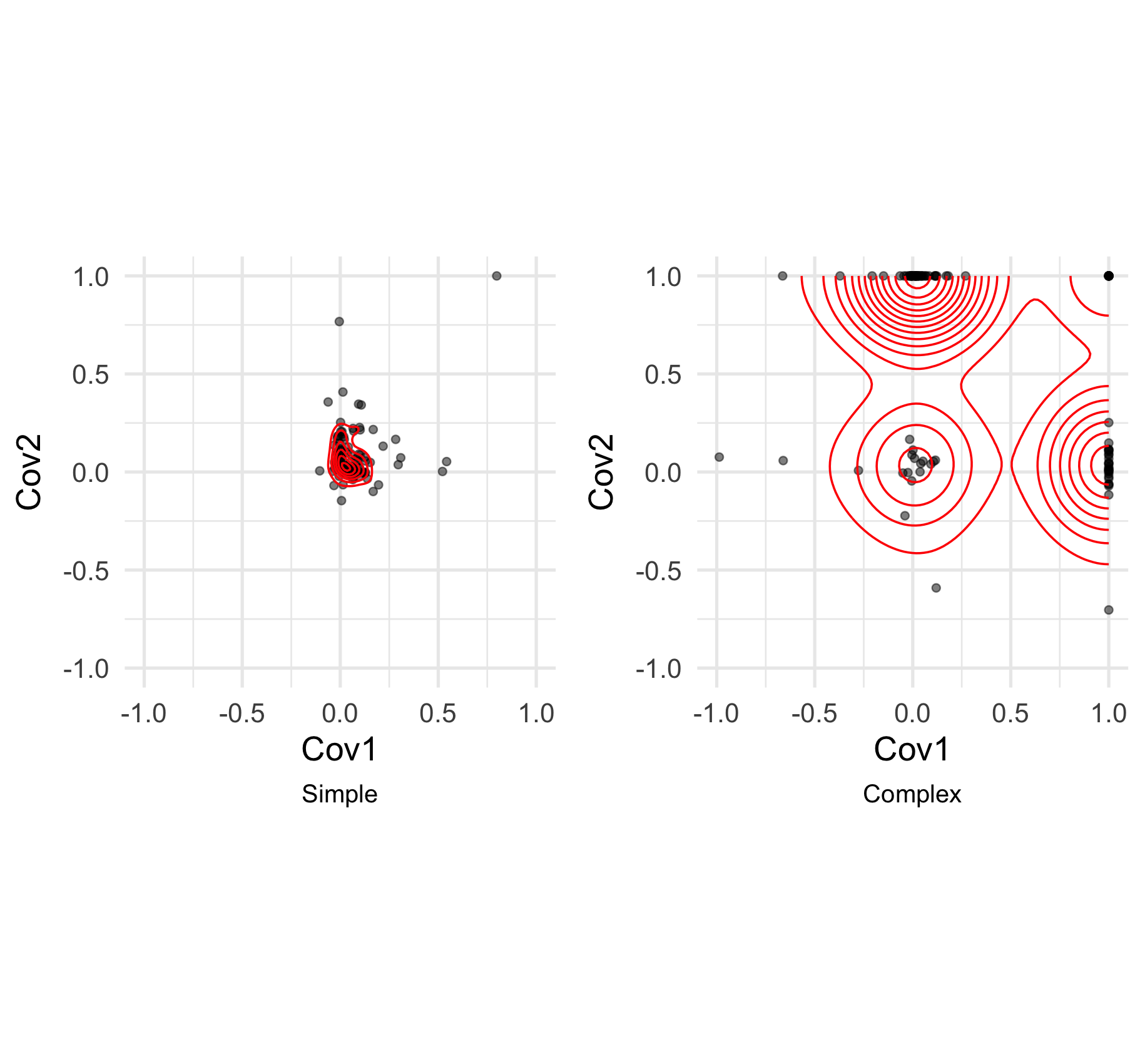}
  \caption{Visualization of heterogeneous subgroup thresholds estimated by the M-learner. Each point represents the threshold identified in a single simulation run. Thresholds greater than 1 or cases where the corresponding variable was not selected are recorded as 1. The red line indicates the estimated density.
Cov1 represnets "X1", Cov2 represnets "X2".}\label{mlearner_border}
\end{figure}

\begin{figure}[H]
  \centering
    \includegraphics[width=0.7\linewidth]{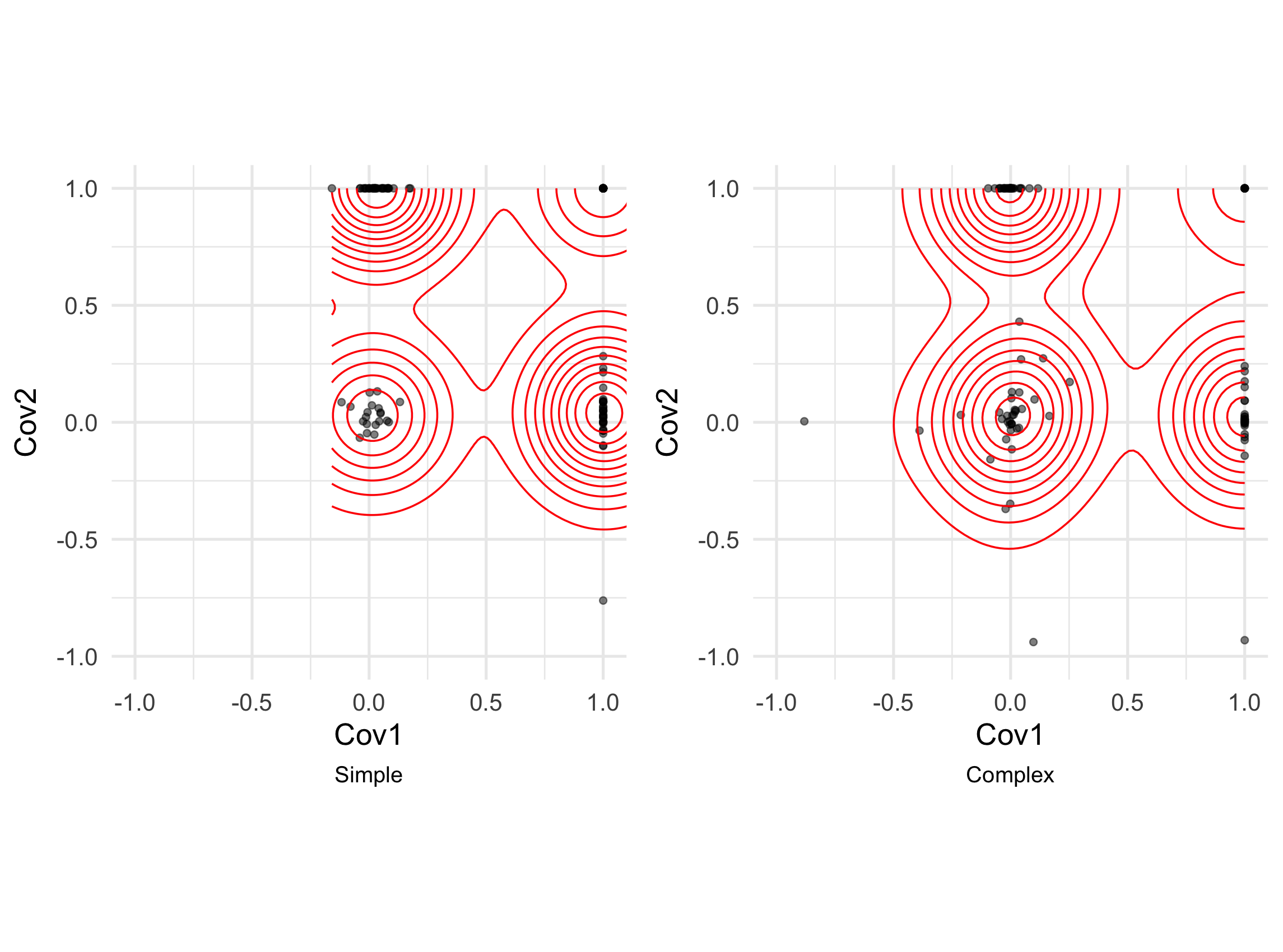}
  \caption{Visualization of heterogeneous subgroup thresholds estimated by the X-learner. Each point represents the threshold identified in a single simulation run. Thresholds greater than 1 or cases where the corresponding variable was not selected are recorded as 1. The red line indicates the estimated density. 
}\label{xleaner_border}
\end{figure}

\begin{figure}[H]
  \centering
    \includegraphics[width=0.7\linewidth]{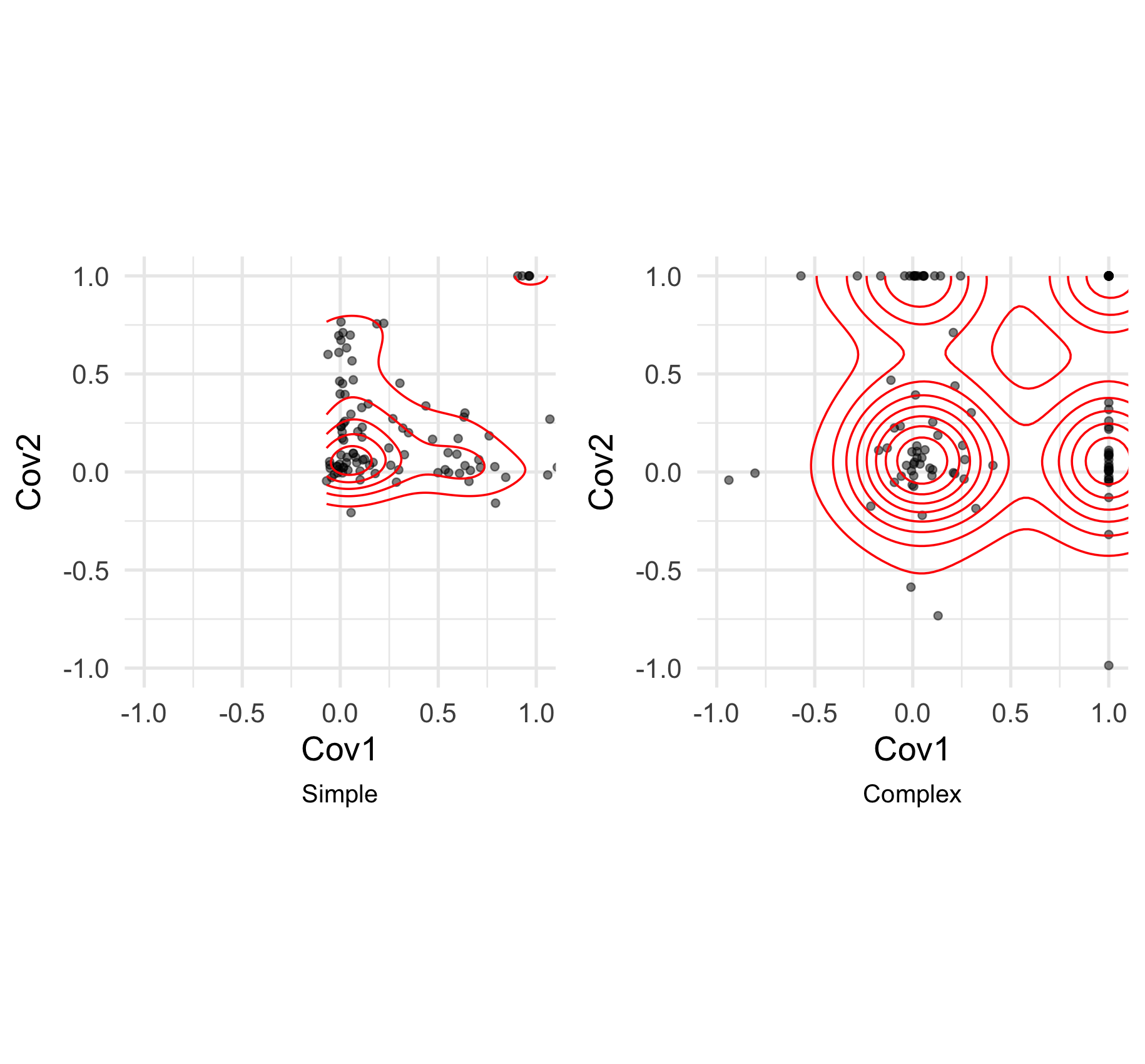}
  \caption{Visualization of heterogeneous subgroup thresholds estimated by the R-learner. Each point represents the threshold identified in a single simulation run. Thresholds greater than 1 or cases where the corresponding variable was not selected are recorded as 1. The red line indicates the estimated density.
}\label{rlearner_border}
\end{figure}

\begin{figure}[H]
  \centering
    \includegraphics[width=0.7\linewidth]{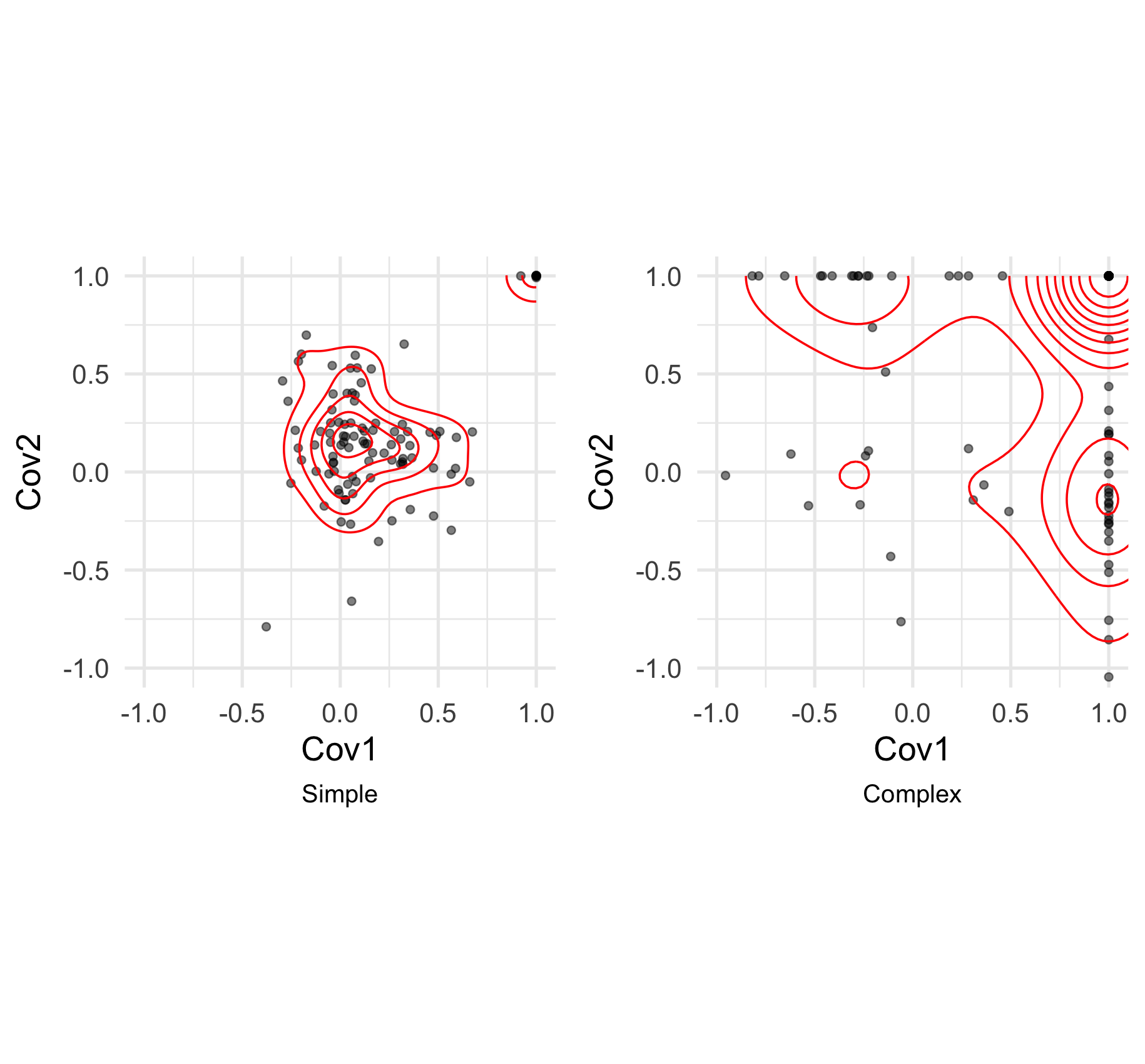}
  \caption{Visualization of heterogeneous subgroup thresholds estimated by the TARNet. Each point represents the threshold identified in a single simulation run. Thresholds greater than 1 or cases where the corresponding variable was not selected are recorded as 1. The red line indicates the estimated density.
}\label{tarnet_border}
\end{figure}

\begin{figure}[H]
  \centering
    \includegraphics[width=0.8\linewidth]{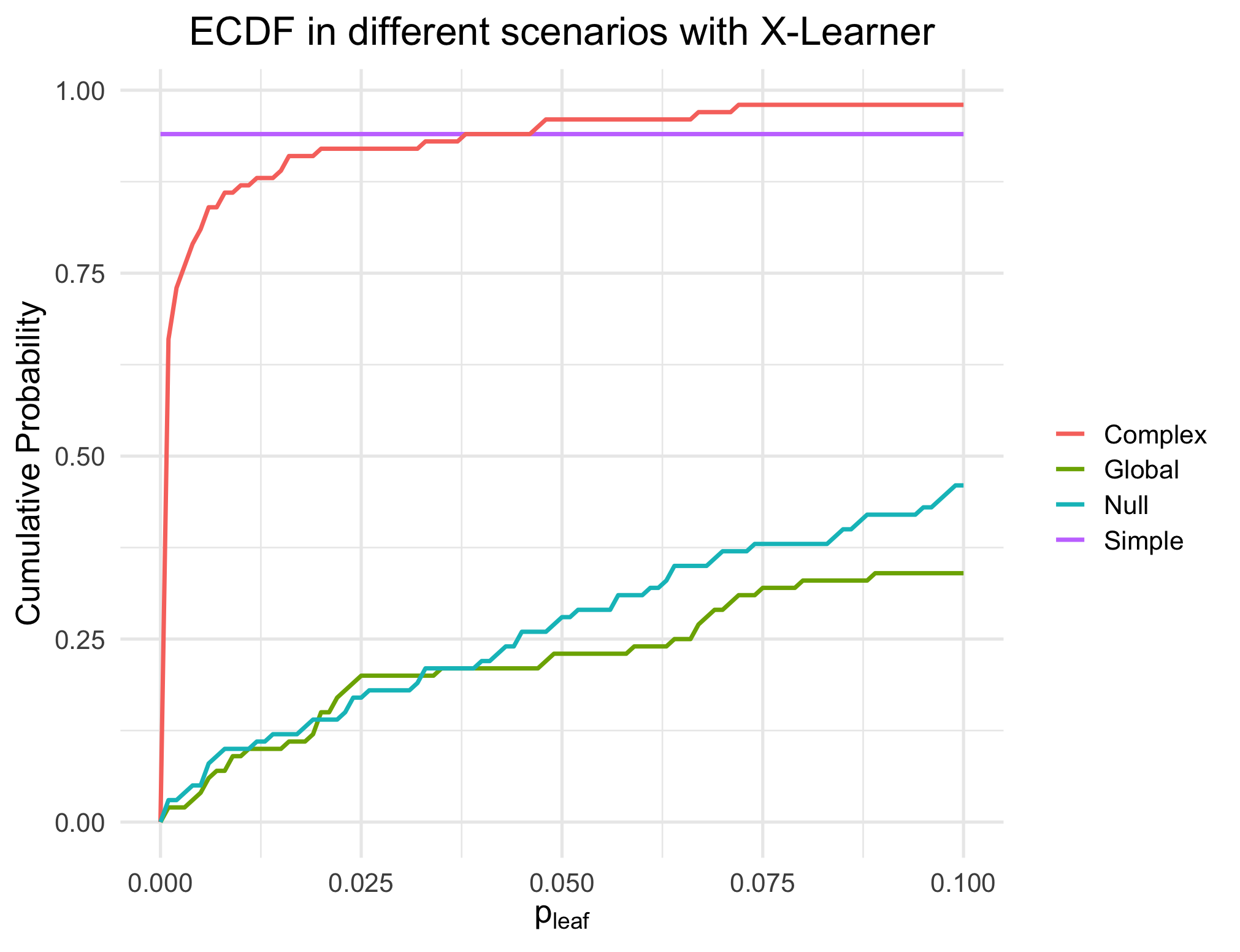}
  \caption{Empirical cumulative distribution functions (ECDF) of \(p_{leaf}\) for X-learner.
}\label{calibration_x}
\end{figure}

\begin{figure}[H]
  \centering
    \includegraphics[width=0.8\linewidth]{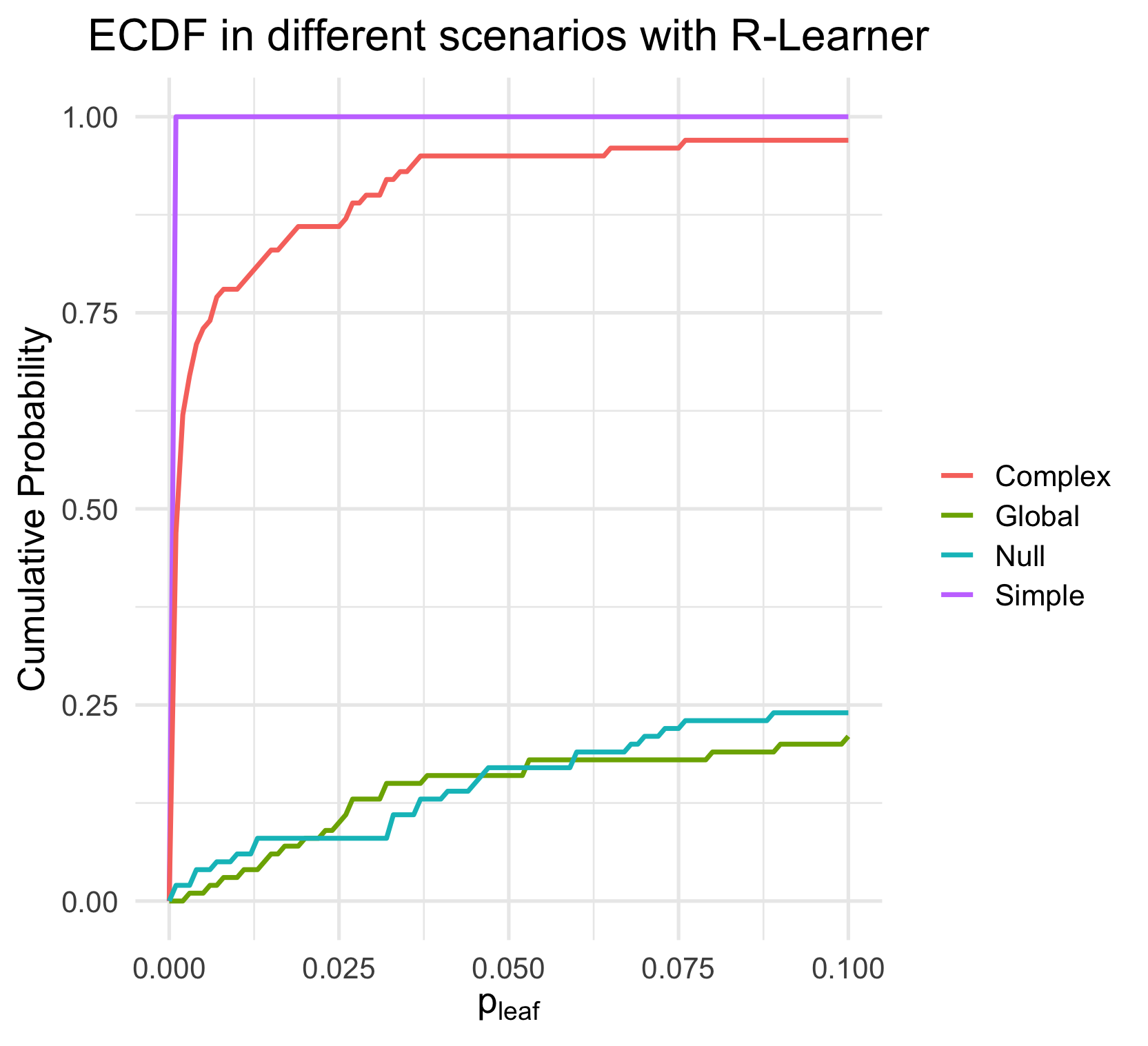}
  \caption{Empirical cumulative distribution functions (ECDF) of \(p_{leaf}\) for R-learner.
}\label{calibration_r}
\end{figure}

\begin{figure}[H]
  \centering
    \includegraphics[width=0.8\linewidth]{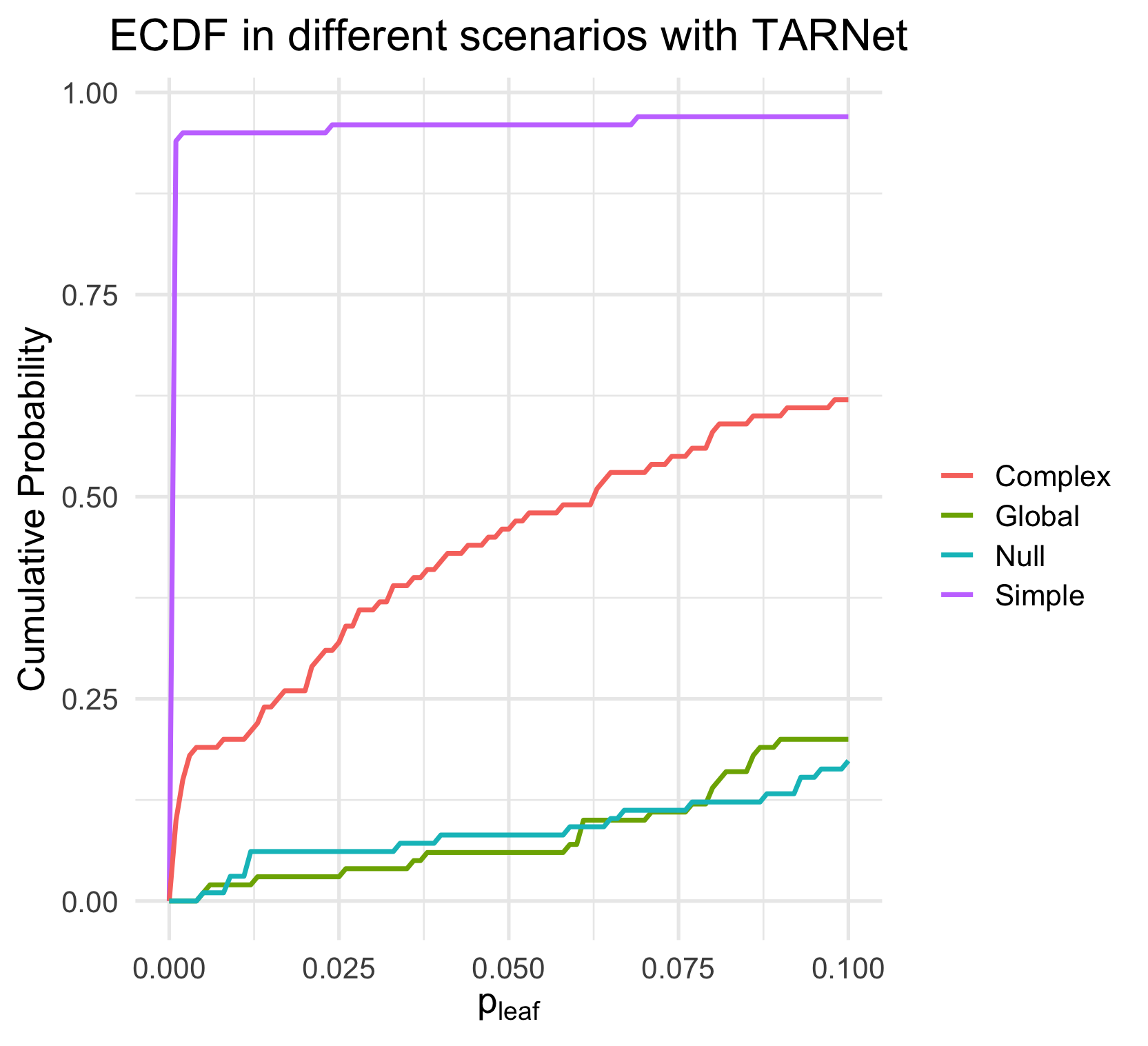}
  \caption{Empirical cumulative distribution functions (ECDF) of \(p_{leaf}\) for TARNet.
}\label{calibration_n}
\end{figure}

To evaluate subgroup identification, we visualized the estimated thresholds defining heterogeneous subgroups across methods. Given that the true subgroup in our simulation is characterized by \(X_1 > 0\) and \(X_2 > 0\), we extracted the estimated boundaries (i.e., a, b such that the identified subgroup is \(X_1 > a, X_2 > b\)). For cases where a method did not select a covariate or the estimated threshold exceeded 1, we set the corresponding value to 1.
The results are shown in Figures \ref{mlearner_border} - \ref{calibration_n}

\subsection{Visualization}
\label{Visualization}
In this section, we visualize an intermediate step of the M-learner algorithm—specifically, the estimation of treatment effects—which provides insights into the underlying mechanisms contributing to its effectiveness.

\subsubsection{Without mediators}
\label{Visualization_no}
When model without mediators, we visualize the estimation of CATTE \(\hat{\tau}^{tot}(x) = \hat{g}_1(x) - \hat{g}_0(x).\)


For each scenario \label{setting_no}, we estimate \(\hat{g}_0(x)\) and \(\hat{g}_1(x)\) using randomized clinical trial data. To facilitate visualization, we construct a synthetic grid of covariates where Cov1 and Cov2 vary from \(-1.5\) to \(1.5\) in increments of \(0.05\), while all other covariates are drawn from a standard normal distribution. The estimated functions are then applied to this grid to compute the CATTE for each unit. We repeat this process across 100 simulated experiments and compute the average CATTE at each grid point. The resulting surface is interpolated using the R package "akima" and visualized to illustrate the spatial patterns in treatment heterogeneity. We visualize the results under a sample size of \(1000\) for each of the four scenarios—Simple, Complex, Global, and Null—using Random Forest and XGBoost as base learners. The visualizations are presented in Appendix Figure \ref{visual_no_1} - \ref{visual_no_4}. 

According to the predefined settings of each scenario in \label{setting_no}, the heterogeneous region in the Simple scenario exhibits positive CATTE values, while the Complex scenario presents negative CATTE values within its heterogeneous region. The Global scenario demonstrates uniformly positive CATTE across the entire covariate space, whereas the Null scenario yields CATTE values consistently close to zero. Our visualizations effectively capture these patterns, providing a clear and accurate reflection of treatment effect heterogeneity across scenarios.

\begin{figure}[H]
  \centering
    \includegraphics[width=0.8\linewidth]{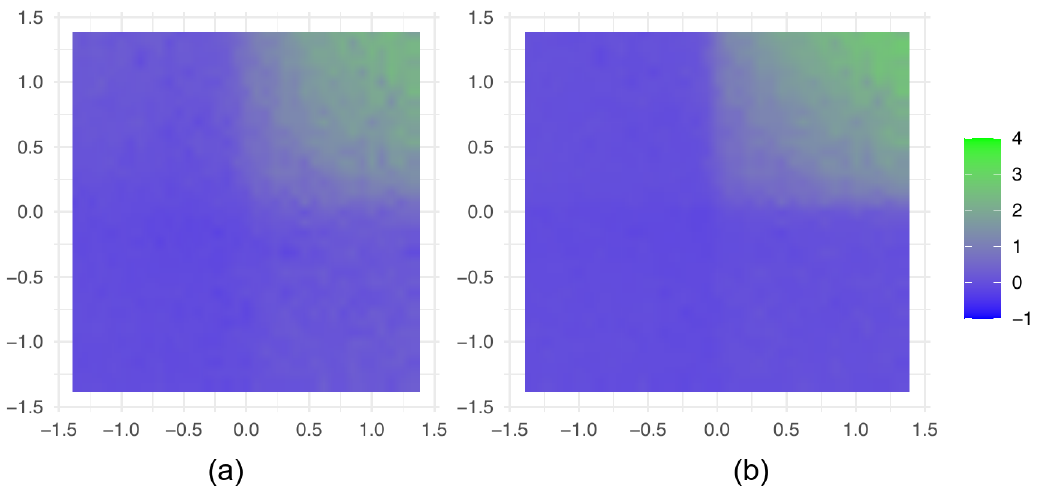}
  \caption{ Visualization of CATTE in the Simple scenario:each panel plots Cov1 (x-axis,\(-1.4\) to \(1.4\)) and Cov2 (y-axis, \(-1.4\) to \(1.4\)), with color representing the estimated CATTE magnitude.(a) visualization of RF learner result, (b) visualization of XGB learner result.
}\label{visual_no_1}
\end{figure}

\begin{figure}[H]
  \centering
    \includegraphics[width=0.8\linewidth]{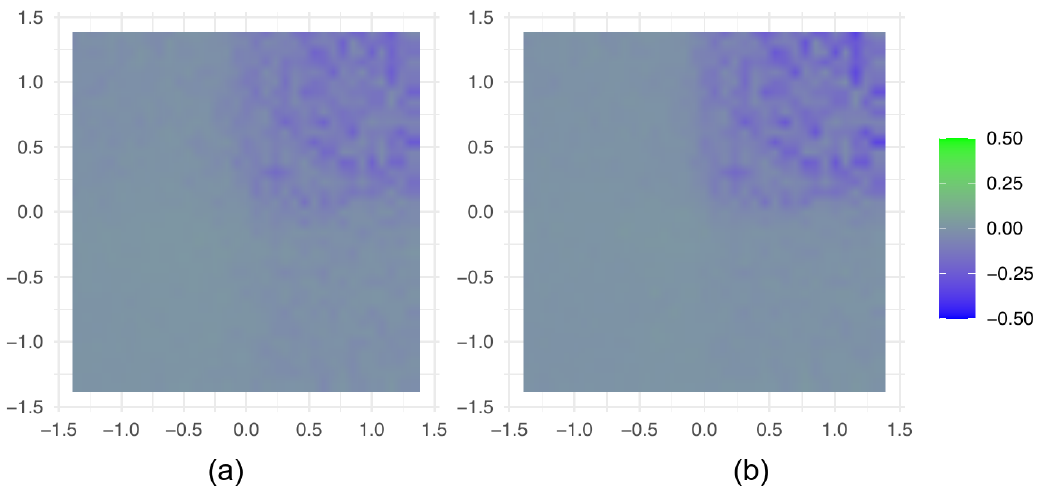}
  \caption{
    Visualization of CATTE in the Complex scenario:each panel plots Cov1 (x-axis,\(-1.4\) to \(1.4\)) and Cov2 (y-axis, \(-1.4\) to \(1.4\)), with color representing the estimated CATTE magnitude.(a) visualization of RF learner result, (b) visualization of XGB learner result.}
  \label{visual_no_2}
\end{figure}

\begin{figure}[h]
  \centering
    \includegraphics[width=0.8\linewidth]{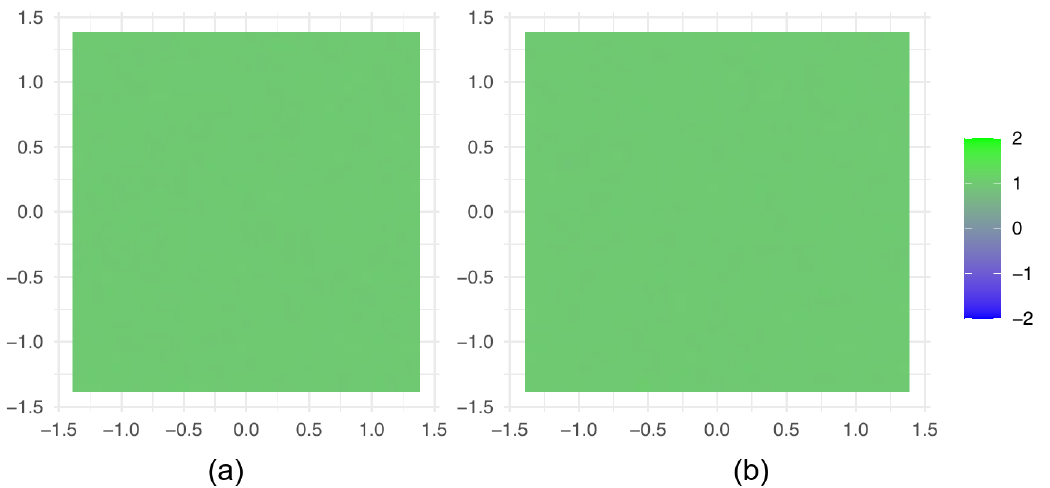}
  \caption{
    Visualization of CATTE in the Global scenario:each panel plots Cov1 (x-axis,\(-1.4\) to \(1.4\)) and Cov2 (y-axis, \(-1.4\) to \(1.4\)), with color representing the estimated CATTE magnitude.(a) visualization of RF learner result, (b) visualization of XGB learner result.}
  \label{visual_no_3}
\end{figure}

\begin{figure}[h]
  \centering
    \includegraphics[width=0.8\linewidth]{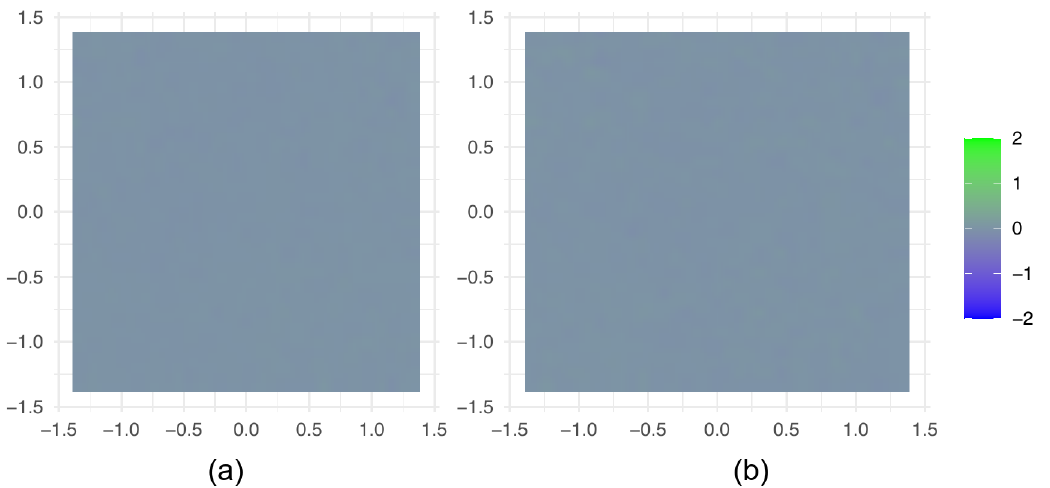}
  \caption{
    Visualization of CATTE in the Null scenario:each panel plots Cov1 (x-axis,\(-1.4\) to \(1.4\)) and Cov2 (y-axis, \(-1.4\) to \(1.4\)), with color representing the estimated CATTE magnitude.(a) visualization of RF learner result, (b) visualization of XGB learner result.}
  \label{visual_no_4}
\end{figure}

\subsubsection{With a mediator}
\label{Visualization_with}
When model with a mediator, we visualize the estimation of CAITE \(
    \hat{\tau}^{ITE}(x) = \hat{g}^{Y}_1(x, \hat{g}^M_1(x)) - \hat{g}^{Y}_1(x, \hat{g}^M_0(x)).\) 

For each scenario \label{setting_with}, we estimate \(\hat{g}^{Y}_1\),\(\hat{g}^M_1(x)\) and \(\hat{g}^M_0(x)\) using randomized clinical trial data. To facilitate visualization, we construct a synthetic grid of covariates where Cov1 and Cov2 vary from \(-1.5\) to \(1.5\) in increments of \(0.05\), while all other covariates are drawn from a standard normal distribution. The estimated functions are then applied to this grid to compute the CAITE for each unit. We repeat this process across 100 simulated experiments and compute the average CAITE at each grid point. The resulting surface is interpolated using the R package "akima" and visualized to illustrate the spatial patterns in treatment heterogeneity. We visualize the results under a sample size of \(1000\) for each of the seven scenarios—Simple-All, Simple-Part, Complex-All, Complex-Part, Simple-Null1, Simple-Null2 and Simple-Global—using Random Forest and XGBoost as base learners. The visualizations are presented in Appendix Figure \ref{visual_1} - \ref{visual_7}. 

According to the predefined settings of each scenario in \label{setting_with}, the heterogeneous region in the Simple-All and Simple-Part scenarios exhibit positive CAITE values, while the Complex-All and Complex-Part scenarios present negative CAITE values within its heterogeneous region. The Simple-Global scenario demonstrates uniformly positive CAITE across the entire covariate space, whereas the Simpe-Null1 and Simple-Null2 scenarios yield CAITE values consistently close to zero. Our visualizations effectively capture these patterns, providing a clear and accurate reflection of treatment effect heterogeneity across scenarios.

\begin{figure}[h]
  \centering
    \includegraphics[width=0.8\linewidth]{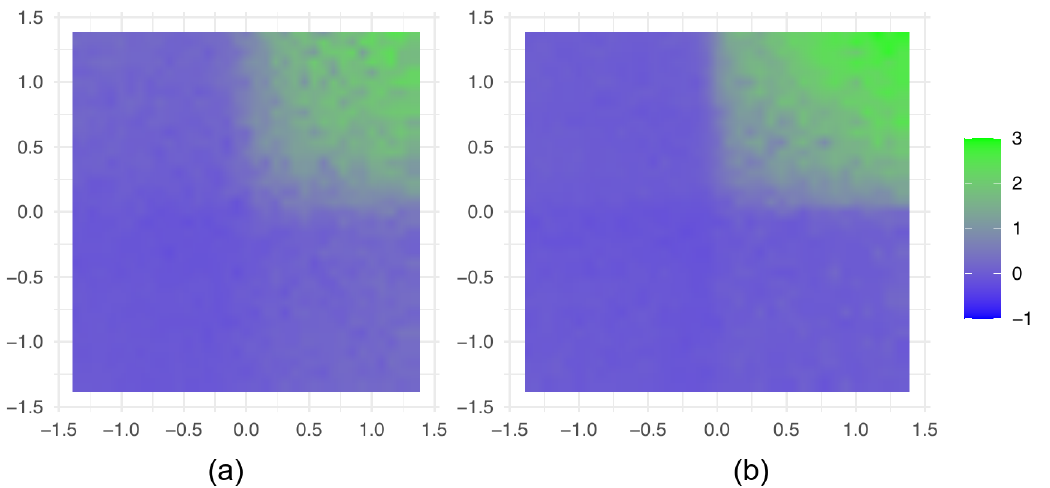}
  \caption{
  Visualization of CAITE in the Simple-All scenario:each panel plots Cov1 (x-axis,\(-1.4\) to \(1.4\)) and Cov2 (y-axis, \(-1.4\) to \(1.4\)), with color representing the estimated CAITE magnitude.(a) visualization of RF learner result, (b) visualization of XGB learner result.}
  \label{visual_1}
\end{figure}

\begin{figure}[h]
  \centering
    \includegraphics[width=0.8\linewidth]{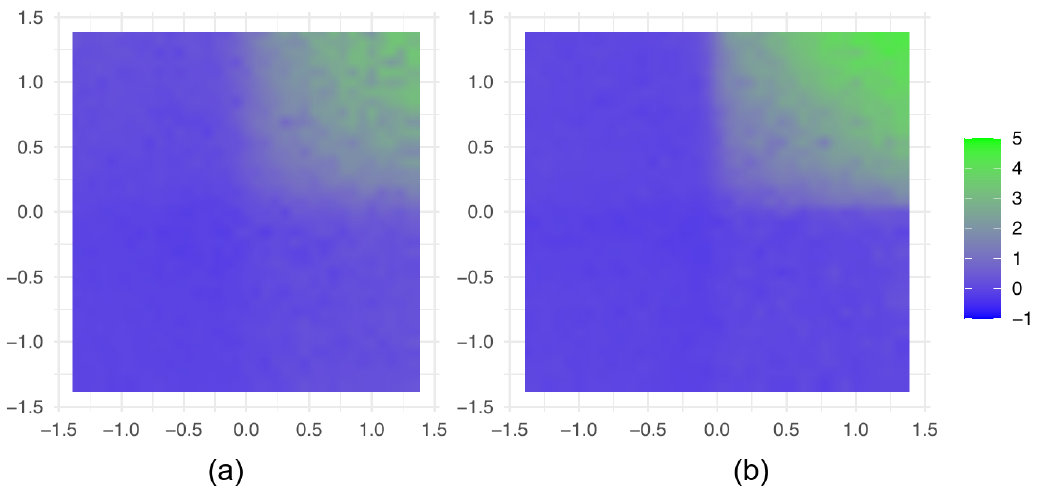}
  \caption{
  Visualization of CAITE in the Simple-Part scenario:each panel plots Cov1 (x-axis,\(-1.4\) to \(1.4\)) and Cov2 (y-axis, \(-1.4\) to \(1.4\)), with color representing the estimated CAITE magnitude.(a) visualization of RF learner result, (b) visualization of XGB learner result.}
  \label{visual_2}
\end{figure}

\begin{figure}[h]
  \centering
    \includegraphics[width=0.8\linewidth]{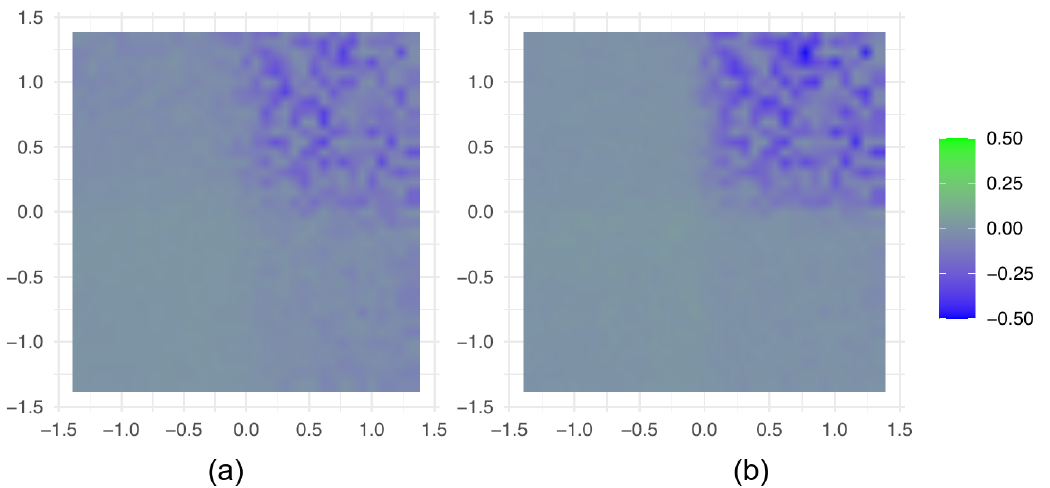}
  \caption{
  Visualization of CAITE in the Complex-All scenario:each panel plots Cov1 (x-axis,\(-1.4\) to \(1.4\)) and Cov2 (y-axis, \(-1.4\) to \(1.4\)), with color representing the estimated CAITE magnitude.(a) visualization of RF learner result, (b) visualization of XGB learner result.}
  \label{visual_3}
\end{figure}

\begin{figure}[h]
  \centering
    \includegraphics[width=0.8\linewidth]{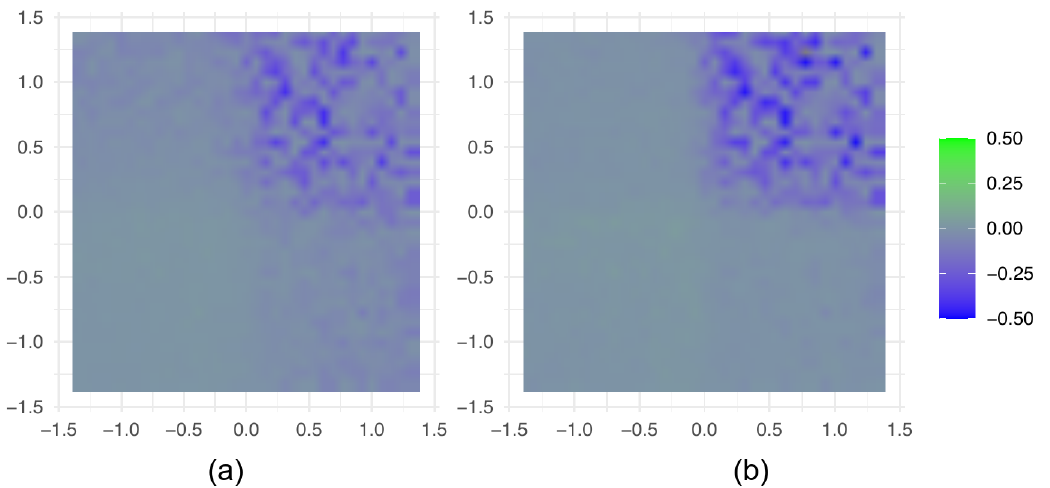}
  \caption{
  Visualization of CAITE in the Complex-Part scenario:each panel plots Cov1 (x-axis,\(-1.4\) to \(1.4\)) and Cov2 (y-axis, \(-1.4\) to \(1.4\)), with color representing the estimated CAITE magnitude.(a) visualization of RF learner result, (b) visualization of XGB learner result.}
  \label{visual_4}
\end{figure}

\begin{figure}[h]
  \centering
    \includegraphics[width=0.8\linewidth]{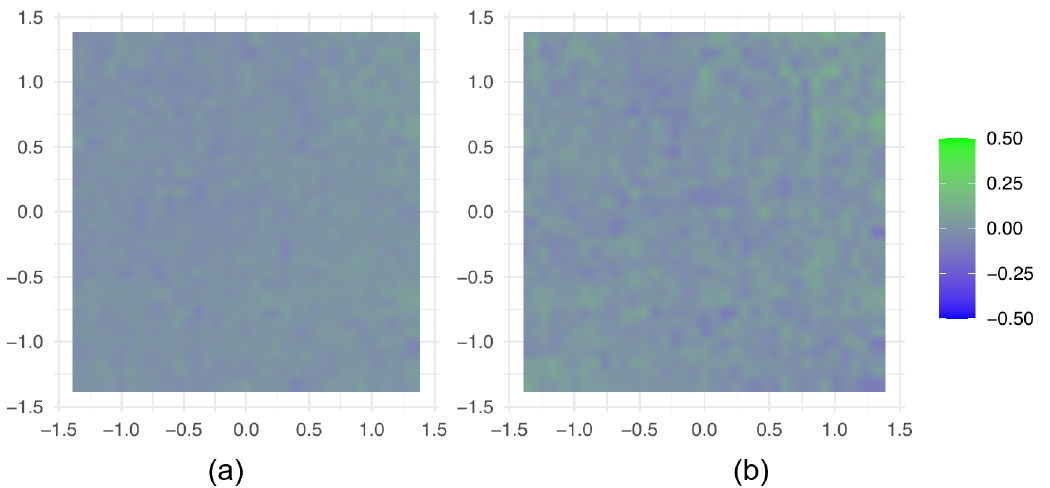}
  \caption{
  Visualization of CAITE in the Simple-Null1 scenario:each panel plots Cov1 (x-axis,\(-1.4\) to \(1.4\)) and Cov2 (y-axis, \(-1.4\) to \(1.4\)), with color representing the estimated CAITE magnitude.(a) visualization of RF learner result, (b) visualization of XGB learner result.}
  \label{visual_5}
\end{figure}

\begin{figure}[h]
  \centering
    \includegraphics[width=0.8\linewidth]{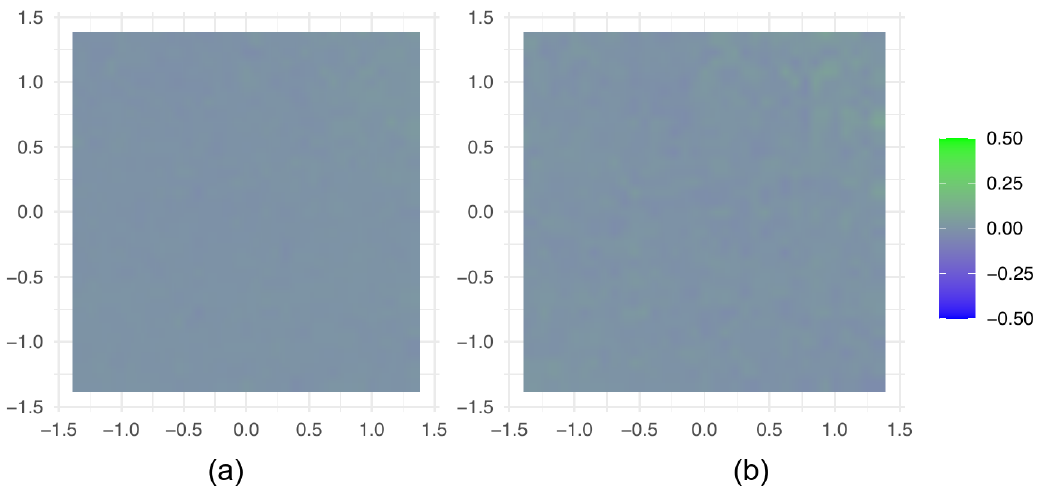}
  \caption{
  Visualization of CAITE in the Simple-Null2 scenario:each panel plots Cov1 (x-axis,\(-1.4\) to \(1.4\)) and Cov2 (y-axis, \(-1.4\) to \(1.4\)), with color representing the estimated CAITE magnitude.(a) visualization of RF learner result, (b) visualization of XGB learner result.}
  \label{visual_6}
\end{figure}

\begin{figure}[h]
  \centering
    \includegraphics[width=0.8\linewidth]{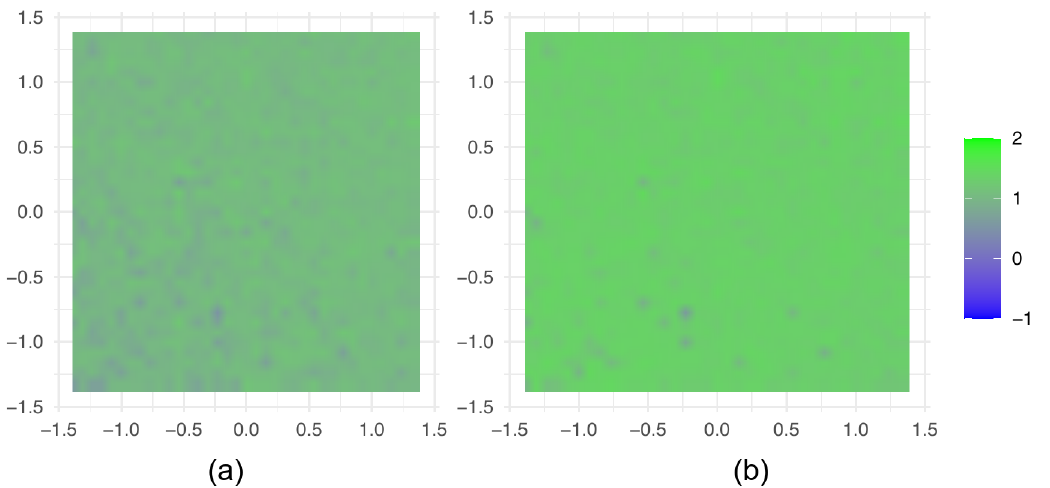}
  \caption{
  Visualization of CAITE in the Simple-Global scenario:each panel plots Cov1 (x-axis,\(-1.4\) to \(1.4\)) and Cov2 (y-axis, \(-1.4\) to \(1.4\)), with color representing the estimated CAITE magnitude.(a) visualization of RF learner result, (b) visualization of XGB learner result.}
  \label{visual_7}
\end{figure}





\end{document}